\documentclass[lettersize,journal]{IEEEtran}
\usepackage{amsmath,amsfonts}
\usepackage[ruled,vlined]{algorithm2e}

\usepackage{array}
\usepackage[caption=false,font=normalsize,labelfont=sf,textfont=sf]{subfig}
\usepackage{textcomp}
\usepackage{stfloats}
\usepackage{url}
\usepackage{verbatim}
\usepackage{graphicx}
\usepackage{cite}

\usepackage{tabularx}
\usepackage{booktabs}       
\usepackage{amsfonts}       
\usepackage{nicefrac}       
\usepackage{microtype}      
\usepackage{xcolor}         
\usepackage{amsmath}
\usepackage{amssymb}
\usepackage{mathtools}
\usepackage{amsthm}

\usepackage{multirow}
\usepackage[table]{xcolor}
\usepackage{color}
\definecolor{mygreen}{RGB}{146, 199, 113} 
\definecolor{taskred}{RGB}{192, 0, 0} 
\definecolor{taskgreen}{RGB}{55, 87, 34} 
\definecolor{taskblue}{RGB}{30, 78, 120} 
\usepackage{graphicx}
\usepackage{adjustbox}
\usepackage{array}
\usepackage{booktabs}
\usepackage{dsfont}
\usepackage{makecell}
\usepackage{wrapfig}
\usepackage{hyperref}
\usepackage{mathrsfs}
\usepackage{xspace}
\usepackage{pifont}
\usepackage{nicematrix}
\usepackage{algpseudocode}

\usepackage[resetlabels]{multibib}
\newcites{supp}{References for Appendix}

\begin{document}

\title{Knowledge Diversion: A Novel Paradigm for Pre-training Decomposable Image Generation Models}

\author{Yucheng Xie,\; Fu Feng,\; Ruixiao Shi,\; Jing Wang$^*$,\; Xin Geng$^*$,~\IEEEmembership{Senior Member,~IEEE},\; Yong Rui,~\IEEEmembership{Fellow,~IEEE}
\thanks{Y. Xie, F. Feng, R. Shi, J. Wang, X. Geng and Y. Rui are with the School of Computer Science and Engineering, Southeast University, Nanjing, China and the Key Laboratory of New Generation Artificial Intelligence Technology and Its Interdisciplinary Applications (Southeast University), Ministry of Education, China 
        (E-mail: \{xieyc, fufeng, eric\_xiao, wangjing91, xgeng\}@seu.edu.cn).}
\thanks{$^*$ Corresponding authors}
}

\markboth{Journal of \LaTeX\ Class Files,~Vol.~14, No.~8, August~2021}%
{Shell \MakeLowercase{\textit{et al.}}: A Sample Article Using IEEEtran.cls for IEEE Journals}


\maketitle

\begin{abstract}
Large-scale pre-trained models have become foundational to modern machine learning, yet conventional pre-training objectives primarily optimize source-domain performance without explicitly considering downstream requirements.
Consequently, transferable and task-specific knowledge remains entangled within a shared parameter space, limiting selective knowledge reuse and efficient downstream adaptation.
This limitation is particularly consequential for image-generation models, which must accommodate diverse conditions and domains.
To address this challenge, we introduce \textit{Knowledge Diversion} (KDiv), a condition-aware pre-training paradigm for learning inherently decomposable generative models. 
Before pre-training, KDiv factorizes randomly initialized weights into rank-one \textit{Knowledge Units}, which are organized into \textit{learngenes} and \textit{tailors}.
During pre-training, continuous condition-aware routing consolidates knowledge shared across conditions in learngenes while directing condition-dependent variations to tailors. 
This structured organization enables learngenes to capture transferable generative priors and tailors to develop condition-specific expertise, allowing the resulting model to be selectively recombined for diverse downstream conditions.
Extensive experiments across class-, text-, and image-conditioned generation demonstrate that KDiv consistently matches or surpasses dense baselines at comparable pre-training cost and achieves the strongest average downstream performance with a trainable parameter budget comparable to low-rank adaptation methods.
\end{abstract} 


\section{Introduction}
\IEEEPARstart{T}{he} continued scaling of neural networks has established large-scale pre-training as a prevailing paradigm for acquiring transferable representations~\cite{qiu2020pre,han2021pre,feng2024redefining}. 
Most approaches, however, follow a ``train-first, adapt-later'' pipeline, optimizing a unified model without explicitly accounting for downstream task structures.
Consequently, transferable and source-specific knowledge remains entangled within the same parameters, limiting model decomposition and task-aware knowledge reuse.
This limitation persists in prevailing adaptation approaches, particularly parameter-efficient fine-tuning (PEFT), where methods such as LoRA~\cite{hulora,hayoulora} and its variants~\cite{zhang2023increlora,valipour2023dylora,liudora} freeze most pre-trained parameters, leaving limited trainable parameters to both suppress source-specific bias and acquire target knowledge under substantial distribution shifts.

\begin{figure}[tb]
  \centering
  \includegraphics[width=\linewidth]{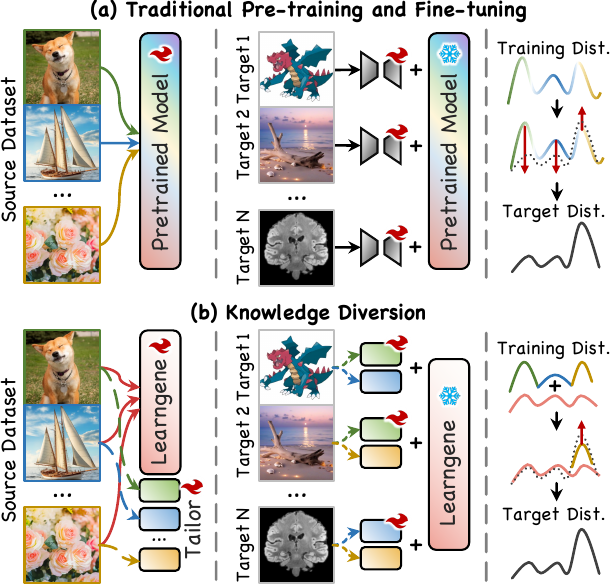}
  \vspace{-0.2in}
  \caption{(a)~Conventional pre-training entangles transferable and source-specific knowledge within a unified backbone, forcing limited downstream updates to both mitigate source-domain bias and capture target-specific variation.
  (b)~KDiv separates condition-agnostic knowledge into shared learngenes and condition-specific knowledge into tailors during pre-training, enabling target models to recombine the shared prior with relevant specialized components. This structured recombination concentrates adaptation on target-specific knowledge and improves alignment with the target distribution.}
  \label{fig:intro}
  \vspace{-0.1in}
\end{figure}

This limitation is particularly consequential for conditional image generation, where models must accommodate diverse conditioning modalities, including class labels, text prompts, and reference images~\cite{peebles2023scalable,rombach2022high,chen2024pixart}.
Although large-scale pre-training provides broad generative priors, the learned knowledge remains closely tied to the source distributions and conditioning schemes, requiring post-hoc adaptation to novel conditions and target domains. 
Under substantial distribution shifts, such knowledge may become redundant~\cite{feng2024transferring}, biased~\cite{ren2024art}, or even detrimental~\cite{wang2019characterizing,rosenstein2005transfer}. 
This raises a fundamental question: \textit{Can we rethink the pre-training process to develop decomposable image-generation models that can be adaptively recombined to meet the specific requirements of downstream tasks?}

Recently, the \textsc{Learngene} framework provides a promising direction by encapsulating transferable knowledge into reusable network fragments, termed \emph{learngenes}, for efficient knowledge transfer and flexible model construction~\cite{feng2023genes}. 
Within this framework, the constraint-based pre-training paradigm~\cite{feng2026constraint} explicitly incorporates cross-scale model initialization into the pre-training objective by encoding size-agnostic knowledge into learngenes through structured parameterizations, including linear transformations~\cite{xia2024transformer}, Tucker decomposition~\cite{xie2026self}, and Kronecker-product factorization~\cite{feng2024wave}.
However, existing approaches consider downstream requirements primarily in terms of model scale for efficient edge deployment, while overlooking variations in task objectives and conditioning modalities.

Accordingly, we introduce \textit{Knowledge Diversion} (KDiv), a novel paradigm that explicitly incorporates downstream task requirements into the pre-training objective to learn inherently decomposable image-generation models\footnote{Knowledge diversion was initially introduced in our conference paper~\cite{xie2025kind}, and the present work extends this earlier study.}.
Rather than encoding all knowledge within an entangled parameterization, KDiv organizes it during pre-training according to condition relevance: transferable, condition-agnostic knowledge is consolidated into \textit{learngenes}, while condition-specific knowledge is routed to \textit{tailors} through continuous soft gating. This structured separation enables adaptive model recombination across diverse conditions and domains, establishing a unified pre-training framework for decomposable class-, text-, and image-conditioned generation.

Specifically, before pre-training, KDiv applies singular value decomposition (SVD) to each randomly initialized principal weight matrix, yielding a set of rank-one \textbf{Knowledge Units}.
Unlike existing methods that factorize pre-trained weights for post-hoc adaptation~\cite{han2023svdiff,zhang2024spectral,robb2020few}, KDiv optimizes these units directly throughout pre-training.
These units are partitioned into \textit{learngenes}, which remain active across conditions to capture condition-agnostic knowledge, and \textit{tailors}, which specialize in condition-specific information.
A modality-specific encoder projects class labels, text prompts, or visual inputs into a unified condition representation, from which a continuous gating network predicts tailor modulation values. 
The shared learngenes and modulated tailors are then combined to reconstruct condition-specific weights, inducing knowledge decomposition during pre-training and enabling adaptive recombination for downstream conditions.

Extensive experiments validate KDiv across multiple model scales and conditioning modalities, including DiT-S/B~\cite{peebles2023scalable} for \textsc{Class-Conditional Generation (C2I)}, PixArt-$\alpha$~\cite{chen2024pixart} for \textsc{Text-Conditional Generation (T2I)}, and PixArt-$\delta$~\cite{chen2024pixartdelta} for \textsc{Image-Conditional Generation (I2I)}.
During pre-training, KDiv matches or surpasses dense baselines in generation quality at comparable cost. For downstream transfer, it achieves the strongest average performance across C2I, T2I, and I2I with a trainable parameter budget comparable to low-rank PEFT baselines, while requiring fewer adaptation steps.
These results indicate that decomposability does not compromise generative capability; instead, separating condition-agnostic from condition-specific knowledge provides an effective inductive bias that improves knowledge organization, transferability, and adaptation efficiency.

Our main contributions are as follows: 
\begin{itemize}
    \item \textbf{Conceptually}, we reformulate generative-model pre-training by shifting its objective from source-domain optimization to downstream-aware knowledge organization.

    \item \textbf{Methodologically}, we propose Knowledge Diversion (KDiv), a condition-aware pre-training framework that explicitly diverts knowledge into condition-agnostic and condition-specific components, yielding inherently decomposable models for adaptive recombination.

    \item \textbf{Empirically}, we establish a comprehensive benchmark spanning C2I, T2I, and I2I, on which KDiv consistently achieves state-of-the-art performance in both pre-training and downstream transfer.
\end{itemize}

\section{Related Work}
\subsection{Traditional Pre-training and Fine-Tuning Paradigm}

Conventional pre-training acquires transferable representations from large-scale data through supervised learning~\cite{dosovitskiy2021an}, contrastive learning~\cite{chen2020simple}, masked prediction~\cite{bao2022beit,he2022masked}, self-distillation~\cite{oquab2024dinov}, or vision--language alignment~\cite{radford2021learning}, followed by task-specific fine-tuning. Despite their diverse objectives, these approaches optimize a unified parameterization in which shared and task-specific knowledge remains entangled, limiting component-level reuse and adaptive recombination. This paradigm also governs generative foundation models: LDM~\cite{rombach2022high} and SDXL~\cite{podell2024sdxl} learn general generative priors during pre-training, while capabilities such as subject personalization and spatial control are introduced through downstream adaptation~\cite{ruiz2023dreambooth,zhang2023adding}. Because condition-agnostic and condition-specific knowledge are not explicitly separated during pre-training, these models lack intrinsic decomposability and rely on post-hoc fine-tuning for new conditions.

To mitigate the cost of full fine-tuning~\cite{touvron2021training,achiam2023gpt}, parameter-efficient fine-tuning (PEFT) restricts updates to prompts~\cite{jia2022visual}, adapters~\cite{houlsby2019parameter,chen2022adaptformer}, low-rank factors~\cite{hulora,hu2023llm}, or spectral components~\cite{han2023svdiff,robb2020few,sun2022singular,zhang2024spectrum,zhang2024spectral}. 
However, because the frozen backbone entangles transferable and source-specific knowledge, these limited trainable parameters must both suppress source-domain bias and capture target-specific variation, constraining adaptation under substantial distribution shifts.

\subsection{Scalable Model Pre-training}

To reduce reliance on post-hoc adaptation, recent work has begun to incorporate downstream model construction into pre-training. A representative direction is the \textit{Learngene} framework~\cite{feng2023genes,feng2024transferring}, which encodes transferable knowledge into reusable network fragments, termed \emph{learngenes}, for initializing models across architectural scales.
Constraint-based pre-training encodes size-agnostic knowledge through structured parameter constraints, including linear transformations~\cite{xia2024transformer}, Kronecker-product factorization~\cite{feng2024wave,feng2026constraint}, and related decomposition strategies~\cite{xie2026fine,xie2026self,shen2026breaking,shen2026unified}. These constraints enable learngenes to be transformed or expanded across model scales, but primarily address architectural scalability rather than heterogeneous downstream task requirements.

Knowledge diversion complements this scale-oriented perspective by organizing pre-trained knowledge according to its relevance to downstream tasks, thereby disentangling broadly reusable knowledge from task-specific information.
Our prior studies progressively established this research direction: KIND~\cite{xie2025kind} first formulated knowledge diversion for class-conditional generation, FAD~\cite{shi2025fad} extended it to frequency-aware downstream adaptation, and DivMorph~\cite{feng2025knowledge} generalized it to cross-morphology transfer in embodied intelligence. 
Building on these foundations, KDiv establishes a unified formulation of knowledge diversion for decomposable pre-training and systematically applies it to image generation models, advancing prior task-specific instantiations into a general mechanism for adaptive decomposition, recombination, and transfer.

\begin{figure*}[tb]
  \centering
  \includegraphics[width=\linewidth]{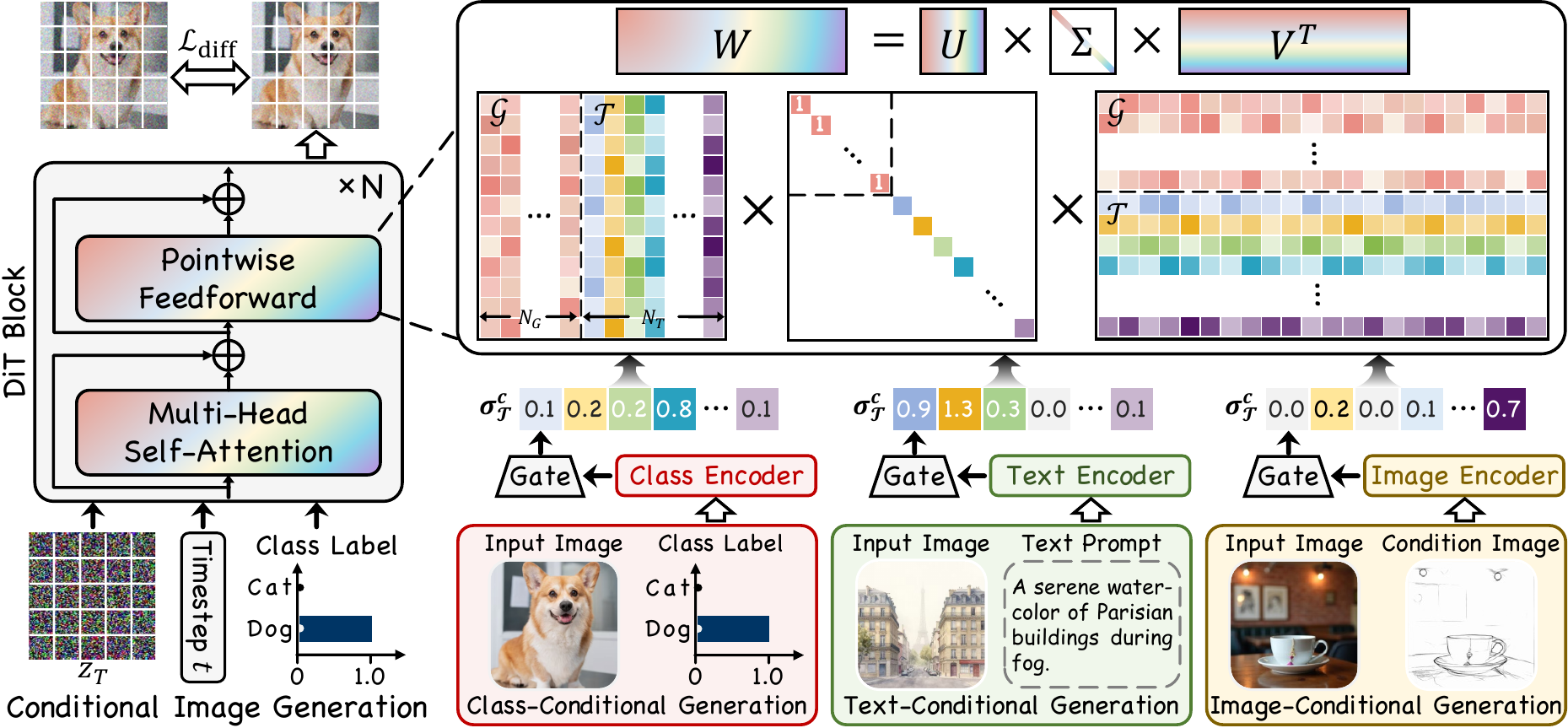}
  \vspace{-0.1in}
  \caption{
  \textbf{Overview of the KDiv framework.}
  KDiv factorizes the principal weight matrices of each DiT block into rank-one Knowledge Units and partitions them into condition-agnostic learngenes and condition-specific tailors.
  Modality-specific encoders and continuous gating predict tailor modulation values, which combine the dynamically weighted tailors with the always-active learngenes to reconstruct condition-specific weights.
  This unified formulation supports class-, text-, and image-conditioned generation under the diffusion training objective.
}
\vspace{-0.1in}
  \label{fig:structure}
\end{figure*}

\section{Methodology}
\subsection{Preliminary}
\subsubsection{Latent Diffusion Models}
Latent diffusion models~\cite{rombach2022high} transfer the diffusion process from the high-resolution pixel space to the latent space by employing an autoencoder $\mathcal{E}$, which encodes an image $x$ into a latent code $z = \mathcal{E}(x)$. A diffusion model is then trained to generate the corresponding latent code in a denoising process, minimizing the following objective:
\begin{equation}
    \mathcal{L}_{\mathrm{diff}}=\mathbb{E}_{z,c,\varepsilon,t}[||\varepsilon-\varepsilon_{\theta}(z_t|c, t)||^{2}_{2}]
\label{eq:loss}
\end{equation}
Here, $\varepsilon_{\theta}$ is a noise prediction network that predicts the noise $\varepsilon$ added to $z_t$ at timestep $t$, conditioned on $c$.

\subsubsection{Diffusion Transformers (DiTs)}
DiT~\cite{peebles2023scalable} replaces the convolutional U-Net denoiser with a Transformer backbone. It divides the latent code $z$ into $P$ patches, projects them into a token sequence $\mathbf{X}\in\mathbb{R}^{P\times D}$, and processes the sequence through $L$ Transformer blocks, each consisting of multi-head self-attention (MSA) and a position-wise feed-forward network (PFF).
Within each MSA module, the input tokens are linearly projected into queries, keys, and values:
\begin{equation*}
\mathbf{Q}=\mathbf{X}W_\mathrm{q},\qquad
\mathbf{K}=\mathbf{X}W_\mathrm{k},\qquad
\mathbf{V}=\mathbf{X}W_\mathrm{v}.
\end{equation*}
Here, $W_\mathrm{q},W_\mathrm{k},W_\mathrm{v}\in\mathbb{R}^{D\times hd}$ are the projection matrices, where $h$ and $d$ denote the number and dimensionality of the attention heads, respectively.
The $i$-th attention head is computed as
\begin{equation}
A_i=
\operatorname{softmax}\left(
\frac{\mathbf{Q}_i\mathbf{K}_i^\top}{\sqrt{d}}
\right)\mathbf{V}_i.
\end{equation}
The head outputs are then concatenated and projected as
\begin{equation}
\operatorname{MSA}(\mathbf{X})
=
\operatorname{Concat}(A_1,\ldots,A_h)W_\mathrm{o},
\label{equ:msa}
\end{equation}
where $W_\mathrm{o}\in\mathbb{R}^{hd\times D}$ is the output projection matrix.

The PFF applies two linear transformations separated by a GELU activation~\cite{hendrycks2016gaussian}:
\begin{equation}
\operatorname{PFF}(\mathbf{X})
=
\operatorname{GELU}(\mathbf{X}W_{\mathrm{in}}+b_1)
W_{\mathrm{out}}+b_2,
\label{equ:mlp}
\end{equation}
where $W_{\mathrm{in}}\in\mathbb{R}^{D\times D'}$ and $W_{\mathrm{out}}\in\mathbb{R}^{D'\times D}$ are the input and output projections, and $D'$ denotes the hidden dimension.

\subsubsection{Problem Formulation}

We consider deriving target models for $T$ heterogeneous downstream tasks from a single pre-trained model $\mathcal{M}_{0}$ with parameters $\Theta_{0}$. Let $\tau_t$ denote the $t$-th downstream task and $\mathcal{M}_t$ the corresponding target model with parameters $\Theta_t$, optimized for its specific objective and data distribution. 
Knowledge Diversion seeks to separate task-agnostic knowledge from task-specific knowledge, enabling a distinct initialization for each downstream task rather than directly reusing the complete pre-trained parameters $\Theta_{0}$.
Formally, we define a task-aware mapping
\begin{equation}
f_{\mathrm{KDiv}}:
\left(\Theta_{0},\tau_t\right)
\mapsto
\Theta_t,
\qquad t=1,\ldots,T,
\label{equ:kdiv_formulation}
\end{equation}
where $f_{\mathrm{KDiv}}$ constructs $\Theta_t$ by preserving transferable task-agnostic knowledge and incorporating task-specific knowledge relevant to $\tau_t$, while excluding information that may impede downstream transfer.

\setlength{\algomargin}{0pt}
\begin{algorithm}[t]
\SetAlCapHSkip{0pt}
\SetInd{0.4em}{0.6em}   
\caption{Knowledge Diversion}
\label{alg:kdiv}

\KwIn{Training dataset $\mathcal{D}$; initial model $\mathcal{M}_0$ with
parameters $\Theta_0=\{W_{\star}^{(l)}\}$; 
numbers of learngenes $N_G$ and tailors $N_T$}

\KwOut{Learngenes $\mathcal{G}$, Tailors $\mathcal{T}$,
and Gating Network $\mathcal{H}$}

\BlankLine
\textbf{Decomposition for Knowledge Unit}\;

\For{$l=1,\ldots,L$}{
    \ForEach{$\star\in\mathcal{S}$}{
        Randomly initialize $W_{\star}^{(l)}$\;
        $\left(U_{\star}^{(l)},\Sigma_{\star}^{(l)},V_{\star}^{(l)}\right)
        \gets \operatorname{SVD}\!\left(W_{\star}^{(l)}\right)$\;
        Construct the rank-one Knowledge Units\par
        \noindent\hfill
        $\displaystyle
        \Theta_{\star}^{(l,i)}
        \gets
        \left(
        \mathbf{u}_{\star}^{(l,i)},
        \mathbf{v}_{\star}^{(l,i)}
        \right),
        \qquad
        i=1,\ldots,r_{\star}
        \mathpunct{;}$
        \hfill\null\par
        Assign $\{\Theta_{\star}^{(l,i)}\}_{i=1}^{N_G}$ to
        $\mathcal{G}$ and
        $\{\Theta_{\star}^{(l,i)}\}_{i=N_G+1}^{r_{\star}}$ to
        $\mathcal{T}$\;
    }
}

\BlankLine
\textbf{Condition-Driven Knowledge Diversion}\;

\While{not converged}{
    Sample an image--condition pair $(x,c)\sim\mathcal{D}$\;
    Compute the condition representation
    $\mathbf{e}_c
    \gets
    \mathcal{E}_{\mathrm{cond}}(c)$\;
    Predict the tailor modulation values
    $\boldsymbol{\sigma}_{\mathcal{T}}^{c}
    \gets
    \mathcal{H}(\mathbf{e}_c)$\;
    
    \For{$l=1,\ldots,L$}{
        \ForEach{$\star\in\mathcal{S}$}{
            Reconstruct the condition-specific weight
            
            \noindent\hfill
            $\displaystyle
            \widetilde{W}_{\star,c}^{(l)}
            \gets
            \mathcal{G}_{\star}^{(l)}
            +
            \sum_{i=1}^{N_T}
            \left[\boldsymbol{\sigma}_{\mathcal{T}}^{c}\right]_i
            \mathcal{T}_{\star,i}^{(l)}
            \mathpunct{;}$
            \hfill\null\par
        }
    }

    Sample $t$ and
    $\boldsymbol{\epsilon}\sim\mathcal{N}(0,I)$, and construct
    $\mathbf{z}_t$\;
    Compute the diffusion objective
    \makebox[\linewidth][c]{%
        $\displaystyle
        \mathcal{L}_{\mathrm{diff}}
        \gets
        \left\|
        \boldsymbol{\epsilon}
        -
        \boldsymbol{\epsilon}_{\widetilde{\Theta}}
        (\mathbf{z}_t\mid c,t)
        \right\|_2^2
        \mathpunct{;}$
    }
    Update $\mathcal{G}$, $\mathcal{T}$, and $\mathcal{H}$
    by minimizing $\mathcal{L}_{\mathrm{diff}}$\;
}

\Return{$\mathcal{G},\mathcal{T},\mathcal{H}$}\;
\end{algorithm}

\subsection{Knowledge Diversion for Decomposable Model Pre-training}
KDiv enables decomposable pretraining by organizing model parameters into Knowledge Units (Sec.~\ref{sec:knowdecom}) and routing condition-agnostic and condition-specific knowledge into distinct components (Sec.~\ref{sec:kdiv}), as summarized in Algorithm~\ref{alg:kdiv}.

\subsubsection{Knowledge Decomposition in Weight Matrices}
\label{sec:knowdecom}
Knowledge diversion requires a structured parameterization that supports fine-grained optimization and routing. 
We therefore apply SVD to randomly initialized weight matrices before pre-training, treating each factor as an independently trainable \textbf{Knowledge Unit} and the basic granularity for selective updates. 
Unlike existing SVD-based methods that factorize pre-trained weights for post-hoc adaptation~\cite{robb2020few,sun2022singular,han2023svdiff}, KDiv learns directly within a decomposed parameter space. Consequently, model decomposability is established during pre-training rather than imposed afterward.

Specifically, for an $L$-layer DiT, let $\Theta$ denote the collection of linear projection matrices in the MSA and PFF modules:
\begin{equation}
\Theta
=
\left\{
W_\star^{(l)}
\;\middle|\;
l=1,\ldots,L,\;
\star \in\mathcal{S}
\right\},
\;
\mathcal{S}
=
\{\mathrm{q},\mathrm{k},\mathrm{v},\mathrm{o},\mathrm{in},\mathrm{out}\},
\end{equation}
Before pre-training, each matrix $W_\star^{(l)}$ is factorized using SVD:
\begin{equation}
W_\star^{(l)}
=
U_\star^{(l)}\Sigma_\star^{(l)}{V_\star^{(l)}}^\top
=
\sum_{i=1}^{r_\star^{(l)}}
\sigma_{\star}^{(l, i)}
\mathbf{u}_{\star}^{(l, i)}
{\mathbf{v}_{\star}^{(l, i)}}^\top,
\label{equ:svd}
\end{equation}
where $r_\star^{(l)}=\operatorname{rank}(W_\star^{(l)})$.
We define each singular-vector pair
$\Theta_\star^{(l,i)}=(\mathbf{u}_\star^{(l,i)},\mathbf{v}_\star^{(l,i)})$
as an independently trainable \textbf{Knowledge Unit}, whose contribution is scaled by $\sigma_\star^{(l,i)}$.

During training, KDiv optimizes individual Knowledge Units rather than dense weight matrices, enabling fine-grained and modular optimization. 
At each forward pass, $W_\star^{(l)}$ is adaptively reconstructed from these units, with the corresponding $\Sigma_\star^{(l)}=\operatorname{diag}\!\left(\sigma_\star^{(l,1)},\ldots,\sigma_\star^{(l,r_\star^{(l)})}\right)$
dynamically predicted by the gating mechanism described in Sec.~\ref{sec:kdiv}.

\subsubsection{Condition-Driven Knowledge Diversion}
\label{sec:kdiv}
Our prior work, KIND~\cite{xie2025kind}, first formulated knowledge diversion for class-conditional generation through discrete class-specific gating. Building on this formulation, KDiv generalizes the paradigm from categorical labels to heterogeneous conditioning modalities, including text prompts and reference images, by introducing a continuous, condition-driven soft-gating mechanism.

Given an input condition $c$, a modality-specific encoder $\mathcal{E}_{\mathrm{cond}}$ maps it to a unified representation,
$\mathbf{e}_c=\mathcal{E}_{\mathrm{cond}}(c)$. 
Depending on the conditioning modality, $\mathcal{E}_{\mathrm{cond}}$ is implemented as a class-embedding layer, text encoder, or image encoder. 
This unified representation enables heterogeneous conditions to be processed by the same knowledge-diversion mechanism.

For each principal weight matrix $W_{\star}^{(l)}$, KDiv partitions its $r_{\star}$ Knowledge Units into $N_G$ \textit{learngenes} and $N_T$ \textit{tailors}, where $r_{\star}=N_G+N_T$. 
The complete sets of learngenes $\mathcal{G}$ and tailors $\mathcal{T}$ across all $L$ Transformer layers are defined as
\begin{equation}
\begin{aligned}
\mathcal{G}
&=
\left\{
\Theta_{\star}^{(l,i)}
\;\middle|\;
l\in[1,L],\;
\star\in\mathcal{S},\;
i\in[1,N_G]
\right\},\\
\mathcal{T}
&=
\left\{
\Theta_{\star}^{(l,i)}
\;\middle|\;
l\in[1,L],\;
\star\in\mathcal{S},\;
i\in[N_G+1,r_{\star}]
\right\}.
\end{aligned}
\label{equ:factor_partition}
\end{equation}
Here, learngenes encode condition-agnostic knowledge, whereas tailors capture condition-specific knowledge.

To induce this functional separation, the gating network $\mathcal{H}$ predicts $\boldsymbol{\sigma}_{\mathcal{T}}^c$ for the $N_T$ tailors conditioned on $\mathbf{e}_c$:
\begin{equation}
\boldsymbol{\sigma}_{\mathcal{T}}^c
=
\mathcal{H}(\mathbf{e}_c)
\in\mathbb{R}^{N_T}.
\label{equ:tailor_coefficients}
\end{equation}
To preserve the condition-agnostic nature of learngenes, their modulation factors are fixed as $\boldsymbol{\sigma}_{\mathcal{G}}=\mathbf{1}_{N_G}$, ensuring identical participation across all conditions. Tailors are instead assigned condition-dependent factors $\boldsymbol{\sigma}_{\mathcal{T}}^{c}$. Together, they form the diagonal modulation matrix $\Sigma_{\star}^{c}=\operatorname{diag}\!\left([\boldsymbol{\sigma}_{\mathcal{G}};\boldsymbol{\sigma}_{\mathcal{T}}^{c}]\right)\in\mathbb{R}^{r_{\star}\times r_{\star}}$.

Accordingly, the condition-specific weight matrix is reconstructed as:
\begin{equation}
\widetilde{W}_{\star,c}^{(l)}
=
\mathcal{G}_{\star}^{(l)}
+
\sum_{i=1}^{N_T}
\left[\boldsymbol{\sigma}_{\mathcal{T}}^{c}\right]_i
\mathcal{T}_{\star}^{(l,i)},
\label{equ:condition_weight}
\end{equation}

The overall training objective is
\begin{equation}
\min_{\mathcal{G},\,\mathcal{T},\,\mathcal{H}}
\mathbb{E}_{(z,c)\sim\mathcal{D}}
\left[
\mathcal{L}_{\mathrm{diff}}
\left(
z,c;\,
\mathcal{G},
\mathcal{T},
\boldsymbol{\sigma}_{\mathcal{T}}^{c}
\right)
\right],
\label{equ:knowledge_diversion}
\end{equation}
where $\mathcal{D}$ is the training distribution and $\mathcal{L}_{\mathrm{diff}}$ denotes the diffusion objective in Eq.~\eqref{eq:loss}. Joint optimization encourages $\mathcal{G}$ to capture condition-agnostic knowledge, while $\mathcal{T}$ specializes through the condition-dependent values predicted by $\mathcal{H}$.


\subsection{Model Recombination for Flexible Adaptation}
\label{sec:recombination}

After knowledge diversion, the pre-trained model is decomposed into condition-agnostic learngenes $\mathcal{G}$ and condition-specific tailors $\mathcal{T}$. Given a target condition set $\mathcal{C}_{\mathrm{tgt}}$, we aggregate the corresponding gating outputs to rank the tailors by relevance and retain the top-$K$. Let $m_K^{(i)}\in\{0,1\}$ indicate whether the $i$-th tailor is selected. The selected tailor set and recombined target model are defined as
\begin{equation}
\mathcal{T}_{\mathrm{tgt}}^{K}
=
\left\{
\Theta^{(i)}\in\mathcal{T}
\mid
m_K^{(i)}=1
\right\},
\qquad
\Theta_{\mathrm{tgt}}
=
\mathcal{G}\cup\mathcal{T}_{\mathrm{tgt}}^{K}.
\label{equ:similar_transfer}
\end{equation}

During downstream adaptation, $\mathcal{G}$ remains frozen and only $\mathcal{T}_{\mathrm{tgt}}^{K}$ is fine-tuned. This preserves the shared generative prior while concentrating optimization on the most relevant condition-specific components, improving both parameter efficiency and robustness to domain shift. See Appendix~\ref{sec:app_recombination} for the detailed recombination and adaptation procedure.

\begin{table*}
\centering
\setlength{\tabcolsep}{0.5 mm}
\caption{
\textbf{Performance of knowledge diversion for training decomposable generative models.}
We compare KDiv with standard training and KIND across class-conditional (C2I), text-conditional (T2I), and image-conditional (I2I) generation tasks. ``Param.'' denotes the model size in terms of parameter count, where values reported as ``activated/total'' correspond to the numbers of activated and total parameters, respectively. ``Time'' denotes the wall-clock time required for model pre-training.
}
\vspace{-0.07in}
\begin{NiceTabular*}{\linewidth}{
@{\extracolsep{\fill}}
ll cc c ccc
@{\hspace{4mm}}
ll cc c ccc
@{}
}
\toprule[1pt]
\multicolumn{16}{c}{\textbf{\textsc{Class-Conditional Generation (C2I)}}} \\
\cmidrule(r){1-8}
\cmidrule{9-16}
\textsc{Model} & \textsc{Train} & \multicolumn{2}{c}{\textsc{Param.}} & \textsc{Time} & FID & sFID & IS & \textsc{Model} & \textsc{Train} & \multicolumn{2}{c}{\textsc{Param.}} & \textsc{Time} & FID & sFID & IS \\
\cmidrule(r){1-8}
\cmidrule{9-16}
\multirow{3}{*}{DiT-S~\cite{peebles2023scalable}} & Standard & \multicolumn{2}{c}{32.96 M} & 8h & 42.85 & 8.57 & 36.06 & \multirow{3}{*}{DiT-B~\cite{peebles2023scalable}} & Standard & \multicolumn{2}{c}{130.51 M} & 19h & 20.33 & 5.97 & 71.71\\
& KIND~\cite{xie2025kind} 
& \multicolumn{2}{c}{32.49$_{\text{ / 60.62}}$ M} 
& 13h & 41.00 & \textbf{8.51} & 38.31 & & KIND~\cite{xie2025kind} 
& \multicolumn{2}{c}{128.91$_{\text{ / 185.14}}$ M} & 26h & 21.03 & \textbf{5.86} & 65.93 \\
& \cellcolor{blue!10}{KDiv} 
& \multicolumn{2}{c}{\cellcolor{blue!10}{\textbf{32.24}$_{\text{ / 34.27}}$ M}} 
& \cellcolor{blue!10}{\textbf{8h}} & \cellcolor{blue!10}{\textbf{40.25}} & \cellcolor{blue!10}{9.38} & \cellcolor{blue!10}{\textbf{38.35}} & & \cellcolor{blue!10}{KDiv} 
& \multicolumn{2}{c}{\cellcolor{blue!10}{\textbf{125.42}$_{\text{ / 133.13}}$ M}} 
& \cellcolor{blue!10}{\textbf{19h}} & \cellcolor{blue!10}{\textbf{19.10}} & \cellcolor{blue!10}{6.07} & \cellcolor{blue!10}{\textbf{75.34}} \\
\bottomrule[1pt]
\end{NiceTabular*}
\par\vspace{1mm}
\begin{NiceTabular*}{\linewidth}{
@{\extracolsep{\fill}}
ll cc c
ccccccc
cccccc
@{}
}
\toprule[1pt]
\multicolumn{18}{c}{\textbf{\textsc{Text-Conditional Generation (T2I)}}} \\
\midrule
\multirow{2}{*}{\textsc{Model}} & \multirow{2}{*}{\textsc{Train}}  & \multicolumn{2}{c}{\multirow{2}{*}{\textsc{Param.}}} & \multirow{2}{*}{\textsc{Time}} & \multicolumn{7}{c}{\textsc{GenEval}} & \multicolumn{6}{c}{\textsc{DPG-Bench}}\\
\cmidrule(r){6-12}
\cmidrule(l){13-18}
& & & & & Single & Two & Count & Colors & Pos. & C-Attr.  & Overall & Global & Entity & Attr. & Rel. & Other & Overall \\
\midrule
\multirow{2}{*}{%
    \shortstack[c]{PixArt-$\alpha$\\[-1pt]
    \cite{chen2024pixart}}%
}
& Standard & \multicolumn{2}{c}{611.64 M} & $\sim$7300h & \textbf{0.972} & 0.510 & 0.475 & 0.824 & 0.083 & 0.100 & 0.494 & \textbf{83.28} & 78.34 & 77.21 & 84.91 & 62.40 & 71.03\\
& \cellcolor{blue!10}{KDiv} 
& \multicolumn{2}{c}{\cellcolor{blue!10}{\textbf{474.53}$_{\text{ / 717.14}}$ M}} 
& \cellcolor{blue!10}{7300+360h} & \cellcolor{blue!10}{0.959} & \cellcolor{blue!10}{\textbf{0.773}} & \cellcolor{blue!10}{\textbf{0.600}} & \cellcolor{blue!10}{\textbf{0.838}} & \cellcolor{blue!10}\textbf{{0.280}} & \cellcolor{blue!10}{\textbf{0.530}} & \cellcolor{blue!10}\textbf{{0.663}} & \cellcolor{blue!10}{81.46} & \cellcolor{blue!10}{\textbf{89.42}} & \cellcolor{blue!10}{\textbf{86.85}} & \cellcolor{blue!10}{\textbf{93.35}} & \cellcolor{blue!10}{\textbf{80.40}} & \cellcolor{blue!10}{\textbf{82.33}} \\
\bottomrule[1pt]
\end{NiceTabular*}
\par\vspace{1mm}

\begin{NiceTabular*}{\linewidth}{
@{\extracolsep{\fill}}
llcc
*{8}{cc}
@{}
}
\toprule[1pt]
\multicolumn{20}{c}{
    \textbf{\textsc{Image-Conditional Generation (I2I)}}
} \\
\midrule
\multirow{2}{*}{\textsc{Model}}
& \multirow{2}{*}{\textsc{Train}}
& \multirow{2}{*}{\textsc{Param.}}
& \multirow{2}{*}{\textsc{Time}}
& \multicolumn{2}{c}{\textsc{BBox}}
& \multicolumn{2}{c}{\textsc{Canny}}
& \multicolumn{2}{c}{\textsc{Depth}}
& \multicolumn{2}{c}{\textsc{HED}}
& \multicolumn{2}{c}{\textsc{Sketch}}
& \multicolumn{2}{c}{\textsc{Normal}}
& \multicolumn{2}{c}{\textsc{Outpaint.}}
& \multicolumn{2}{c}{\textsc{Segment.}} \\
\cmidrule(lr){5-6}
\cmidrule(lr){7-8}
\cmidrule(lr){9-10}
\cmidrule(lr){11-12}
\cmidrule(lr){13-14}
\cmidrule(lr){15-16}
\cmidrule(lr){17-18}
\cmidrule(l){19-20}
& & & 
& LPIPS & CLIP
& LPIPS & CLIP
& LPIPS & CLIP
& LPIPS & CLIP
& LPIPS & CLIP
& LPIPS & CLIP
& LPIPS & CLIP
& LPIPS & CLIP \\
\midrule

\multirow{2}{*}{%
    \shortstack[c]{PixArt-$\delta$\\[-1pt]
    \cite{chen2024pixartdelta}}%
}
& Standard
& 294.92 M
& 81h
& 0.248 & 88.99
& 0.285 & 93.41
& 0.234 & 93.41
& 0.277 & 94.01
& 0.253 & 91.64
& 0.401 & 90.15
& 0.052 & 94.78
& 0.456 & 88.34 \\

& \cellcolor{blue!10}{KDiv}
& \cellcolor{blue!10}{\textbf{294.92}$_{\text{ / 468.03}}$ M}
& \cellcolor{blue!10}{\textbf{79h}}
& \cellcolor{blue!10}{\textbf{0.231}}
& \cellcolor{blue!10}{\textbf{90.55}}
& \cellcolor{blue!10}{\textbf{0.272}}
& \cellcolor{blue!10}{\textbf{94.11}}
& \cellcolor{blue!10}{\textbf{0.223}}
& \cellcolor{blue!10}{\textbf{93.78}}
& \cellcolor{blue!10}{\textbf{0.256}}
& \cellcolor{blue!10}{\textbf{94.63}}
& \cellcolor{blue!10}{\textbf{0.239}}
& \cellcolor{blue!10}{\textbf{92.34}}
& \cellcolor{blue!10}{\textbf{0.388}}
& \cellcolor{blue!10}{\textbf{90.78}}
& \cellcolor{blue!10}{\textbf{0.052}}
& \cellcolor{blue!10}{\textbf{94.81}}
& \cellcolor{blue!10}{\textbf{0.441}}
& \cellcolor{blue!10}{\textbf{88.90}} \\

\bottomrule[1pt]
\end{NiceTabular*}
\vspace{-0.1in}
\label{tab:main_kdiv}
\end{table*}

\section{Experimental Setup}
\subsection{Datasets}

We evaluate KDiv across three conditional generation settings. For \textsc{Class-Conditional Generation (C2I)}, models are pretrained on ImageNet-1K~\cite{deng2009imagenet} and transferred to eight downstream datasets covering fine-grained categories, natural scenes, and remote-sensing imagery, including CUB-200, Food-101, and RESISC45. For \textsc{Text-Conditional Generation (T2I)}, knowledge diversion is performed on Fine-T2I~\cite{ma2026finet2i}, with downstream evaluation spanning diverse visual domains such as Pokémon, LEGO, and Signatures. For \textsc{Image-Conditional Generation (I2I)}, KDiv is trained on eight base conditions from Subjects200K~\cite{tan2024ominicontrol} and evaluated on novel controls from COCO2017~\cite{lin2014microsoft}, following CtrLoRA~\cite{xu2024ctrlora}. Collectively, these datasets provide a unified benchmark for evaluating generation quality and downstream transferability across categorical, textual, and visual conditions. Complete dataset specifications and processing details are provided in Appendix~\ref{sec:app_dataset}.

\subsection{Baselines}
\subsubsection{Baselines for Knowledge Diversion}

We compare KDiv with standard dense training under matched backbone architectures. 
For C2I, experiments are conducted on DiT-S/B~\cite{peebles2023scalable}, with KIND~\cite{xie2025kind} included as an earlier knowledge-diversion method based on discrete class-specific routing. 
For T2I and I2I, we use PixArt-$\alpha$~\cite{chen2024pixart} and PixArt-$\delta$~\cite{chen2024pixartdelta}, respectively. 
These controlled comparisons isolate the effect of knowledge diversion across categorical, textual, and visual conditioning.

\subsubsection{Baselines for Downstream Adaptation}

For downstream adaptation, we compare KDiv with representative PEFT methods, including SVDiff~\cite{han2023svdiff}, OFT~\cite{qiu2023controlling}, LoRA~\cite{hulora}, PiSSA~\cite{meng2024pissa}, LoHa~\cite{yeh2024navigating}, and DoRA~\cite{liudora}. 
These baselines span spectral, orthogonal, and low-rank adaptation, enabling systematic evaluation of transferability, distribution-shift robustness, and parameter efficiency. See Appendix~\ref{sec:app_baseline} for more details.

\subsection{Evaluation}
C2I pre-training is evaluated on ImageNet using Fr\'echet Inception Distance (FID$\downarrow$)~\cite{heusel2017gans}, spatial FID (sFID$\downarrow$)~\cite{nash2021generating}, and Inception Score (IS$\uparrow$)~\cite{salimans2016improved}. For domain-shifted C2I transfer, we report Fr\'echet DINOv2 Distance (FDD$\downarrow$), Kernel DINOv2 Distance (KDD$\downarrow$)~\cite{stein2023exposing}, and CLIP Maximum Mean Discrepancy (CMMD$\downarrow$)~\cite{jayasumana2024rethinking}. T2I pre-training is assessed using GenEval$\uparrow$~\cite{ghosh2023geneval} and DPG-Bench$\uparrow$~\cite{hu2024ella}, while I2I pre-training is evaluated on held-out pairs using LPIPS$\downarrow$~\cite{zhang2018unreasonable} and CLIP-I$\uparrow$~\cite{radford2021learning}. Downstream T2I and I2I performance is measured using LPIPS$\downarrow$, SSIM$\uparrow$~\cite{wang2004image}, and CLIP-T/CLIP-I$\uparrow$, capturing perceptual fidelity, structural similarity, and semantic alignment, respectively. 
See Appendix~\ref{sec:hyper} for more training details.

\begin{figure*}[tb]
\centering
\includegraphics[width=\linewidth]{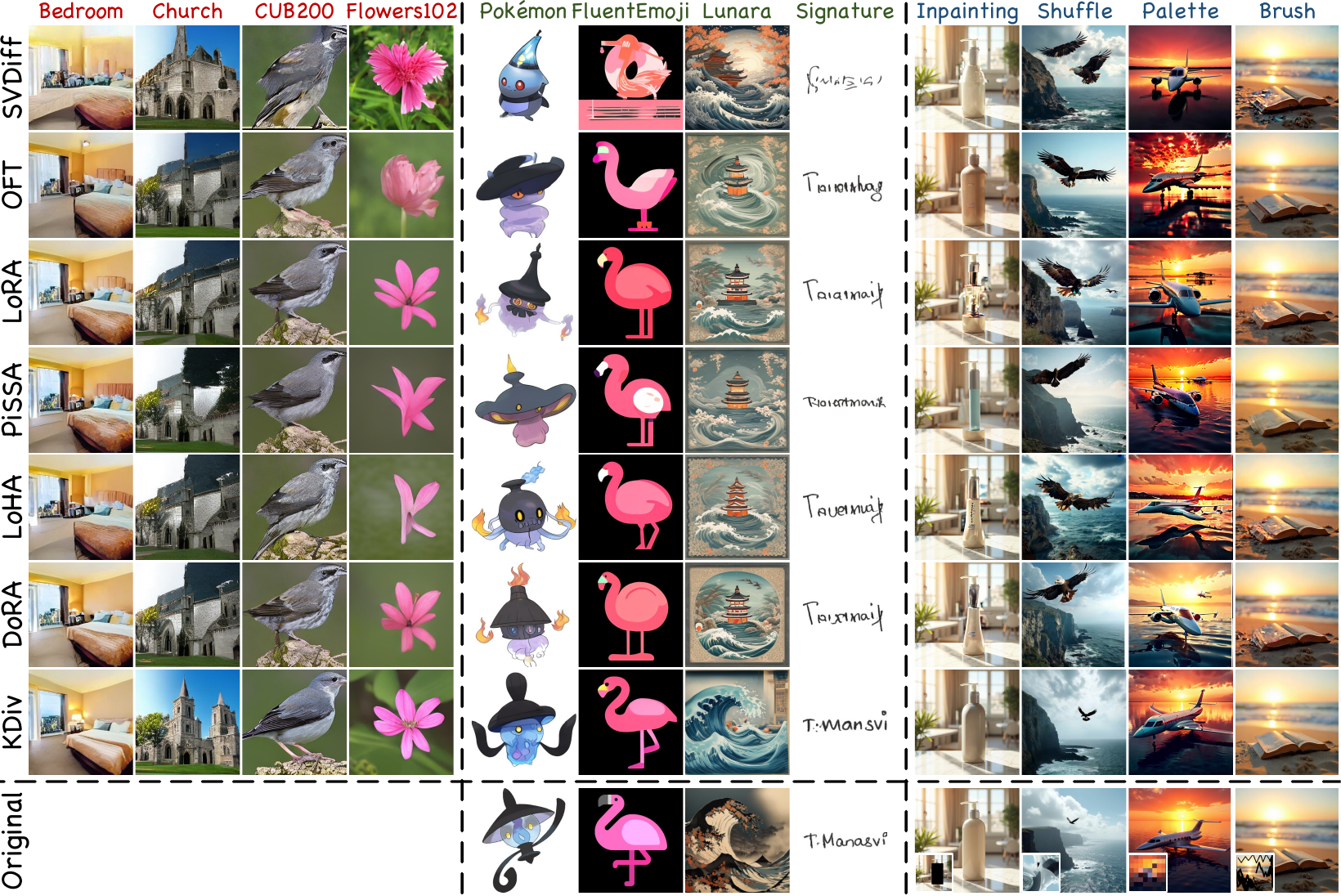}
\vspace{-0.2in}
\caption{
\textbf{Qualitative comparison on novel downstream conditions.}
Representative samples generated by KDiv and competing PEFT methods for \textcolor{taskred}{\textsc{Class-Conditional Generation (C2I)}},
\textcolor{taskgreen}{\textsc{Text-Conditional Generation (T2I)}}, and
\textcolor{taskblue}{\textsc{Image-Conditional Generation (I2I)}} tasks. 
Additional qualitative comparisons are provided in Appendix Figs.~\ref{fig:app_down_C2I_T2I} and~\ref{fig:app_down_I2I}.
}
\label{fig:downstream}
\end{figure*}

\renewcommand{\arraystretch}{0.91}
\begin{table*}
\centering
\setlength{\tabcolsep}{0.8 mm}
\caption{\textbf{Performance of knowledge diversion in adapting to novel conditions.}
We compare KDiv with parameter-efficient adaptation methods on previously unseen downstream conditions. Performance is evaluated using FDD~($\downarrow$), KDD~($\downarrow$), and CMMD~($\downarrow$) for C2I, and LPIPS~($\downarrow$), SSIM~($\uparrow$), and CLIP-T/CLIP-I~($\uparrow$) for T2I/I2I, respectively. ``GPU Time'' denotes the adaptation cost, and ``Param.'' denotes the average number of trainable parameters.}
\vspace{-0.06in}
\resizebox{\linewidth}{!}{
\begin{tabular}{@{}lcccccccccccc|ccc@{}}
\toprule[1pt]
\multicolumn{16}{c}{\textbf{Class-Conditional Generation (C2I)}} \\
\midrule
& \multicolumn{3}{c}{\textsc{CUB-200}} & \multicolumn{3}{c}{\textsc{DF20M}} & \multicolumn{3}{c}{\textsc{Flowers-102}} & \multicolumn{3}{c|}{\textsc{Food-101}} & \multicolumn{3}{c}{\textsc{Cost}}\\ 
\cmidrule(lr){2-4} 
\cmidrule(lr){5-7}
\cmidrule(lr){8-10}
\cmidrule(lr){11-13} 
\cmidrule(l){14-16} 
& FDD$\downarrow$ & KDD$\downarrow$ & CMMD$\downarrow$ & FDD$\downarrow$ & KDD$\downarrow$ & CMMD$\downarrow$ & FDD$\downarrow$ & KDD$\downarrow$ & CMMD$\downarrow$ & FDD$\downarrow$ & KDD$\downarrow$ & CMMD$\downarrow$ & \multicolumn{2}{c}{\textsc{GPU Time}} & \textsc{Param.} \\ 
\cmidrule(r){1-13}
\cmidrule(l){14-16}
SVDiff~\cite{han2023svdiff}
& 1987 & 3.775 & 1.597
& 1878 & 6.332 & 2.378
& 2083 & 3.510 & 1.587
& 1594 & 3.832 & 2.690
& \multicolumn{2}{c}{0.92 h} & 0.02M \\
OFT~\cite{qiu2023controlling}
& 1130 & 1.452 & 0.635
& 977 & 3.213 & 1.366
& 1651 & 1.992 & 1.175
& 1290 & 2.979 & 2.217
& \multicolumn{2}{c}{1.26 h} & 15.03M \\
LoRA~\cite{hulora}
& 1095 & 1.313 & 0.535
& 1053 & 3.544 & 1.387
& 1575 & 1.674 & 0.889
& 1245 & 2.835 & 2.007
& \multicolumn{2}{c}{1.13 h} & 11.80M \\
PiSSA~\cite{meng2024pissa}
& 1051 & 1.230 & 0.531
& 930 & 3.089 & 1.263
& 1596 & 1.767 & 0.973
& 1208 & 2.697 & 1.964
& \multicolumn{2}{c}{1.14 h} & 11.80M \\
LoHa~\cite{yeh2024navigating}
& 1313 & 1.819 & 0.704
& 1262 & 4.274 & 1.663
& 1649 & 1.920 & 0.877
& 1349 & 3.107 & 2.216
& \multicolumn{2}{c}{1.85 h} & 11.80M \\
DoRA~\cite{liudora}
& 1077 & 1.286 & 0.557
& 1037 & 3.516 & 1.377
& 1553 & 1.652 & 0.959
& 1245 & 2.865 & 1.992
& \multicolumn{2}{c}{1.38 h} & 11.82M \\
\cmidrule(r){1-13} 
\cmidrule(l){14-16} 
KIND~\cite{xie2025kind}
& 1053 & 1.246 & 0.568
& 803 & 2.682 & 1.120
& \textbf{1421} & \textbf{1.453} & \textbf{0.683}
& 1177 & 2.655 & 1.817
& \multicolumn{2}{c}{1.22 h} & 11.80M \\
\cellcolor{blue!10}{KDiv}
& \cellcolor{blue!10}{\textbf{995}} & \cellcolor{blue!10}{\textbf{1.098}} & \cellcolor{blue!10}{\textbf{0.386}}
& \cellcolor{blue!10}{\textbf{698}} & \cellcolor{blue!10}{\textbf{2.306}} & \cellcolor{blue!10}{\textbf{0.863}}
& \cellcolor{blue!10}{1507} & \cellcolor{blue!10}{1.576} & \cellcolor{blue!10}{0.685}
& \cellcolor{blue!10}{\textbf{1075}} & \cellcolor{blue!10}{\textbf{2.341}} & \cellcolor{blue!10}{\textbf{1.568}}
& \multicolumn{2}{c}{\cellcolor{blue!10}{0.90 h}} & \cellcolor{blue!10}{11.80M} \\
\midrule[1pt]

& \multicolumn{3}{c}{\textsc{LSUN Bedroom}} & \multicolumn{3}{c}{\textsc{LSUN Church}} & \multicolumn{3}{c}{\textsc{RESISC45}} & \multicolumn{3}{c|}{\textsc{Stanford Cars}} & \multicolumn{3}{c}{\textsc{Average}} \\ 
\cmidrule(lr){2-4} 
\cmidrule(lr){5-7}
\cmidrule(lr){8-10}
\cmidrule(lr){11-13} 
\cmidrule(l){14-16} 
& FDD$\downarrow$ & KDD$\downarrow$ & CMMD$\downarrow$ & FDD$\downarrow$ & KDD$\downarrow$ & CMMD$\downarrow$ & FDD$\downarrow$ & KDD$\downarrow$ & CMMD$\downarrow$ & FDD$\downarrow$ & KDD$\downarrow$ & CMMD$\downarrow$ & FDD$\downarrow$ & KDD$\downarrow$ & CMMD$\downarrow$\\ 
\cmidrule(r){1-13}
\cmidrule(l){14-16}
SVDiff~\cite{han2023svdiff}
& 2001 & 5.661 & 3.419
& 1711 & 4.378 & 3.472
& 1489 & 5.765 & 2.115
& 2012 & 6.088 & 2.621
& 1844 & 4.917 & 2.485 \\
OFT~\cite{qiu2023controlling}
& 1282 & 4.064 & 2.389
& 1304 & 3.116 & 2.598
& 913 & 3.573 & 1.711
& 730 & 2.029 & 0.918
& 1160 & 2.802 & 1.626 \\
LoRA~\cite{hulora}
& 1285 & 4.035 & 2.287
& 1301 & 3.129 & 2.605
& 703 & 2.962 & 1.503
& 692 & 1.755 & 0.899
& 1119 & 2.656 & 1.514 \\
PiSSA~\cite{meng2024pissa}
& 1202 & 3.806 & 2.181
& 1241 & 2.908 & 2.468
& 658 & 2.611 & 1.465
& 615 & 1.535 & 0.796
& 1063 & 2.455 & 1.455 \\
LoHa~\cite{yeh2024navigating}
& 1496 & 4.559 & 2.574
& 1406 & 3.479 & 2.854
& 826 & 3.485 & 1.598
& 908 & 2.514 & 1.275
& 1276 & 3.145 & 1.720 \\
DoRA~\cite{liudora}
& 1279 & 4.045 & 2.290
& 1315 & 3.193 & 2.616
& 699 & 2.940 & 1.499
& 708 & 1.896 & 0.871
& 1114 & 2.674 & 1.520 \\
\cmidrule(r){1-13} 
\cmidrule(l){14-16} 
KIND~\cite{xie2025kind}
& 1101 & 3.660 & 1.959
& 1190 & 2.897 & 2.236
& \textbf{566} & \textbf{2.319} & \textbf{1.384}
& \textbf{554} & \textbf{1.386} & 0.622
& 983 & 2.287 & 1.299 \\
\cellcolor{blue!10}{KDiv}
& \cellcolor{blue!10}{\textbf{1052}} & \cellcolor{blue!10}{\textbf{3.616}} & \cellcolor{blue!10}{\textbf{1.881}}
& \cellcolor{blue!10}{\textbf{1130}} & \cellcolor{blue!10}{\textbf{2.701}} & \cellcolor{blue!10}{\textbf{2.057}}
& \cellcolor{blue!10}{665} & \cellcolor{blue!10}{2.746} & \cellcolor{blue!10}{1.470}
& \cellcolor{blue!10}{604} & \cellcolor{blue!10}{1.530} & \cellcolor{blue!10}{\textbf{0.608}}
& \cellcolor{blue!10}{\textbf{966}} & \cellcolor{blue!10}{\textbf{2.239}} & \cellcolor{blue!10}{\textbf{1.190}} \\
\bottomrule[1pt]
\toprule[1pt]
\multicolumn{16}{c}{\textbf{Text-Conditional Generation (T2I)}} \\
\midrule
& \multicolumn{3}{c}{\textsc{Adventurers}} & \multicolumn{3}{c}{\textsc{FluentEmoji}} & \multicolumn{3}{c}{\textsc{Hearthstone}} & \multicolumn{3}{c|}{\textsc{LEGO}} & \multicolumn{3}{c}{\textsc{Cost}}\\ 
\cmidrule(lr){2-4} 
\cmidrule(lr){5-7}
\cmidrule(lr){8-10}
\cmidrule(lr){11-13} 
\cmidrule(l){14-16} 
& LPIPS$\downarrow$ & SSIM$\uparrow$ & CLIP-T$\uparrow$ & LPIPS$\downarrow$ & SSIM$\uparrow$ & CLIP-T$\uparrow$ & LPIPS$\downarrow$ & SSIM$\uparrow$ & CLIP-T$\uparrow$ & LPIPS$\downarrow$ & SSIM$\uparrow$ & CLIP-T$\uparrow$ & \multicolumn{2}{c}{\textsc{GPU Time}} & \textsc{Param.} \\ 
\cmidrule(r){1-13}
\cmidrule(l){14-16}
SVDiff~\cite{han2023svdiff}
& 0.646 & 0.343 & 26.48
& 0.661 & 0.216 & 19.92
& 0.644 & 0.112 & 23.04
& 0.692 & 0.271 & 21.19
& \multicolumn{2}{c}{0.19 h} & 0.23M \\
OFT~\cite{qiu2023controlling}
& 0.616 & 0.386 & 27.45
& 0.399 & 0.579 & 20.24
& 0.490 & 0.282 & 24.57
& 0.658 & 0.298 & 20.38
& \multicolumn{2}{c}{0.33 h} & 92.82M \\
LoRA~\cite{hulora}
& 0.629 & 0.369 & 27.70
& 0.441 & 0.544 & 20.08
& 0.468 & 0.302 & 24.84
& 0.670 & 0.300 & 22.20
& \multicolumn{2}{c}{0.26 h} & 64.51M \\
PiSSA~\cite{meng2024pissa}
& 0.622 & 0.385 & 27.41
& 0.410 & 0.545 & 20.38
& 0.490 & 0.277 & 24.56
& 0.661 & 0.318 & 21.13
& \multicolumn{2}{c}{0.25 h} & 64.51M \\
LoHa~\cite{yeh2024navigating}
& 0.629 & 0.363 & 27.27
& 0.448 & 0.541 & 19.82
& 0.481 & 0.287 & 24.70
& 0.655 & 0.316 & \textbf{22.69}
& \multicolumn{2}{c}{0.30 h} & 64.51M \\
DoRA~\cite{liudora}
& 0.631 & 0.375 & 27.71
& 0.444 & 0.535 & 19.85
& 0.484 & 0.279 & 24.80
& 0.668 & 0.308 & 21.89
& \multicolumn{2}{c}{0.39 h} & 64.52M \\
\cmidrule(r){1-13} 
\cmidrule(l){14-16} 
\cellcolor{blue!10}{KDiv}
& \cellcolor{blue!10}{\textbf{0.608}} & \cellcolor{blue!10}{\textbf{0.411}} & \cellcolor{blue!10}{\textbf{27.92}}
& \cellcolor{blue!10}{\textbf{0.379}} & \cellcolor{blue!10}{\textbf{0.585}} & \cellcolor{blue!10}{\textbf{20.86}}
& \cellcolor{blue!10}{\textbf{0.454}} & \cellcolor{blue!10}{\textbf{0.321}} & \cellcolor{blue!10}{\textbf{25.21}}
& \cellcolor{blue!10}{\textbf{0.648}} & \cellcolor{blue!10}{\textbf{0.318}} & \cellcolor{blue!10}{21.91}
& \multicolumn{2}{c}{\cellcolor{blue!10}{0.23 h}} & \cellcolor{blue!10}{64.52M} \\
\midrule[1pt]
& \multicolumn{3}{c}{\textsc{Lunara}} & \multicolumn{3}{c}{\textsc{Pokémon}} & \multicolumn{3}{c}{\textsc{RickMorty}} & \multicolumn{3}{c|}{\textsc{Signatures}} & \multicolumn{3}{c}{\textsc{Average}} \\ 
\cmidrule(lr){2-4} 
\cmidrule(lr){5-7}
\cmidrule(lr){8-10}
\cmidrule(lr){11-13} 
\cmidrule(l){14-16} 
& LPIPS$\downarrow$ & SSIM$\uparrow$ & CLIP-T$\uparrow$ & LPIPS$\downarrow$ & SSIM$\uparrow$ & CLIP-T$\uparrow$ & LPIPS$\downarrow$ & SSIM$\uparrow$ & CLIP-T$\uparrow$ & LPIPS$\downarrow$ & SSIM$\uparrow$ & CLIP-T$\uparrow$ & LPIPS$\downarrow$ & SSIM$\uparrow$ & CLIP-T$\uparrow$ \\ 
\cmidrule(r){1-13}
\cmidrule(l){14-16}
SVDiff~\cite{han2023svdiff}
& 0.692 & 0.324 & 26.56
& 0.603 & 0.401 & 24.91
& 0.809 & 0.196 & 22.45
& 0.298 & 0.738 & 21.19
& 0.631 & 0.325 & 23.22 \\
OFT~\cite{qiu2023controlling}
& 0.657 & 0.353 & 27.20
& 0.522 & 0.483 & 26.87
& 0.787 & 0.243 & 25.25
& 0.183 & 0.842 & 22.58
& 0.539 & 0.433 & 24.32 \\
LoRA~\cite{hulora}
& 0.661 & 0.337 & 27.01
& 0.518 & 0.471 & 26.90
& 0.790 & 0.203 & 25.17
& 0.178 & 0.848 & 22.30
& 0.544 & 0.422 & 24.52 \\
PiSSA~\cite{meng2024pissa}
& 0.654 & \textbf{0.359} & 27.20
& 0.518 & 0.488 & 26.77
& 0.779 & 0.241 & 25.64
& 0.160 & 0.859 & 22.25
& 0.537 & 0.434 & 24.42 \\
LoHa~\cite{yeh2024navigating}
& 0.662 & 0.343 & 27.15
& 0.532 & 0.471 & 26.94
& 0.794 & 0.190 & 24.60
& 0.177 & 0.847 & 22.54
& 0.547 & 0.420 & 24.46 \\
DoRA~\cite{liudora}
& 0.663 & 0.329 & 26.82
& 0.514 & 0.481 & 27.00
& 0.788 & 0.210 & 25.34
& 0.178 & 0.846 & 22.78
& 0.546 & 0.420 & 24.52 \\
\cmidrule(r){1-13} 
\cmidrule(l){14-16} 
\cellcolor{blue!10}{KDiv}
& \cellcolor{blue!10}{\textbf{0.641}} & \cellcolor{blue!10}{0.346} & \cellcolor{blue!10}{\textbf{28.01}}
& \cellcolor{blue!10}{\textbf{0.494}} & \cellcolor{blue!10}{\textbf{0.499}} & \cellcolor{blue!10}{\textbf{28.05}}
& \cellcolor{blue!10}{\textbf{0.772}} & \cellcolor{blue!10}{\textbf{0.256}} & \cellcolor{blue!10}{\textbf{25.64}}
& \cellcolor{blue!10}{\textbf{0.156}} & \cellcolor{blue!10}{\textbf{0.866}} & \cellcolor{blue!10}{\textbf{23.96}}
& \cellcolor{blue!10}{\textbf{0.519}} & \cellcolor{blue!10}{\textbf{0.450}} & \cellcolor{blue!10}{\textbf{25.20}} \\
\bottomrule[1pt]
\toprule[1pt]
\multicolumn{16}{c}{\textbf{Image-Conditional Generation (I2I)}} \\
\midrule
& \multicolumn{3}{c}{\textsc{Blur}} & \multicolumn{3}{c}{\textsc{Brush}} & \multicolumn{3}{c}{\textsc{Grayscale}} & \multicolumn{3}{c|}{\textsc{Inpainting}} & \multicolumn{3}{c}{\textsc{Cost}}\\ 
\cmidrule(lr){2-4} 
\cmidrule(lr){5-7}
\cmidrule(lr){8-10}
\cmidrule(lr){11-13} 
\cmidrule(l){14-16} 
& LPIPS$\downarrow$ & SSIM$\uparrow$ & CLIP-I$\uparrow$ & LPIPS$\downarrow$ & SSIM$\uparrow$ & CLIP-I$\uparrow$ & LPIPS$\downarrow$ & SSIM$\uparrow$ & CLIP-I$\uparrow$ & LPIPS$\downarrow$ & SSIM$\uparrow$ & CLIP-I$\uparrow$ & \multicolumn{2}{c}{\textsc{GPU Time}} & \textsc{Param.} \\ 
\cmidrule(r){1-13}
\cmidrule(l){14-16}
SVDiff~\cite{han2023svdiff}
& 0.339 & 0.807 & 89.47
& 0.138 & 0.799 & 95.87
& 0.232 & 0.679 & 95.96
& 0.143 & 0.784 & 96.78
& \multicolumn{2}{c}{0.30 h} & 0.15M \\
OFT~\cite{qiu2023controlling}
& 0.161 & 0.934 & 95.36
& 0.158 & 0.782 & 94.29
& 0.195 & 0.712 & 97.21
& 0.154 & 0.773 & 96.68
& \multicolumn{2}{c}{0.40 h} & 18.59M \\
LoRA~\cite{hulora}
& 0.173 & 0.933 & 94.86
& 0.140 & 0.798 & 95.74
& 0.203 & 0.718 & 96.92
& 0.159 & 0.765 & 95.44
& \multicolumn{2}{c}{0.31 h} & 14.02M \\
PiSSA~\cite{meng2024pissa}
& 0.168 & 0.934 & 95.33
& 0.152 & 0.787 & 94.84
& 0.199 & 0.718 & 97.16
& 0.160 & 0.766 & 95.28
& \multicolumn{2}{c}{0.29 h} & 14.02M \\
LoHa~\cite{yeh2024navigating}
& 0.234 & 0.899 & 92.73
& 0.137 & 0.800 & 95.85
& 0.203 & \textbf{0.724} & 96.90
& 0.141 & 0.785 & 96.76
& \multicolumn{2}{c}{0.41 h} & 14.02M \\
DoRA~\cite{liudora}
& 0.195 & 0.923 & 94.28
& 0.141 & 0.793 & 95.50
& 0.203 & 0.723 & 96.86
& 0.139 & \textbf{0.789} & 96.87
& \multicolumn{2}{c}{0.46 h} & 14.21M \\
\cmidrule(r){1-13} 
\cmidrule(l){14-16} 
\cellcolor{blue!10}{KDiv}
& \cellcolor{blue!10}{\textbf{0.142}} & \cellcolor{blue!10}{\textbf{0.946}} & \cellcolor{blue!10}{\textbf{95.71}}
& \cellcolor{blue!10}{\textbf{0.128}} & \cellcolor{blue!10}{\textbf{0.809}} & \cellcolor{blue!10}{\textbf{96.10}}
& \cellcolor{blue!10}{\textbf{0.191}} & \cellcolor{blue!10}{0.717} & \cellcolor{blue!10}{\textbf{97.43}}
& \cellcolor{blue!10}{\textbf{0.131}} & \cellcolor{blue!10}{0.787} & \cellcolor{blue!10}{\textbf{97.31}}
& \multicolumn{2}{c}{\cellcolor{blue!10}{0.27 h}} & \cellcolor{blue!10}{14.02M} \\
\midrule[1pt]

& \multicolumn{3}{c}{\textsc{Jpeg}} & \multicolumn{3}{c}{\textsc{Palette}} & \multicolumn{3}{c}{\textsc{Pixel}} & \multicolumn{3}{c|}{\textsc{Shuffle}} & \multicolumn{3}{c}{\textsc{Average}} \\ 
\cmidrule(lr){2-4} 
\cmidrule(lr){5-7}
\cmidrule(lr){8-10}
\cmidrule(lr){11-13} 
\cmidrule(l){14-16} 
& LPIPS$\downarrow$ & SSIM$\uparrow$ & CLIP-I$\uparrow$ & LPIPS$\downarrow$ & SSIM$\uparrow$ & CLIP-I$\uparrow$ & LPIPS$\downarrow$ & SSIM$\uparrow$ & CLIP-I$\uparrow$ & LPIPS$\downarrow$ & SSIM$\uparrow$ & CLIP-I$\uparrow$ & LPIPS$\downarrow$ & SSIM$\uparrow$ & CLIP-I$\uparrow$ \\ 
\cmidrule(r){1-13}
\cmidrule(l){14-16}
SVDiff~\cite{han2023svdiff}
& 0.393 & 0.583 & 89.43
& 0.387 & 0.589 & 81.45
& 0.464 & 0.540 & 86.46
& 0.687 & \textbf{0.249} & 83.66
& 0.348 & 0.629 & 89.88 \\
OFT~\cite{qiu2023controlling}
& \textbf{0.302} & 0.663 & 93.41
& 0.234 & 0.726 & 88.33
& 0.372 & 0.600 & 89.62
& 0.678 & 0.249 & 84.45
& 0.282 & 0.680 & 92.42 \\
LoRA~\cite{hulora}
& 0.333 & 0.635 & 91.90
& 0.241 & 0.725 & 88.24
& 0.405 & 0.593 & 89.39
& 0.678 & 0.238 & 84.22
& 0.292 & 0.676 & 92.09 \\
PiSSA~\cite{meng2024pissa}
& 0.311 & 0.667 & 93.37
& 0.226 & 0.732 & 88.84
& 0.398 & 0.573 & 88.80
& 0.679 & 0.240 & \textbf{84.72}
& 0.287 & 0.677 & 92.29 \\
LoHa~\cite{yeh2024navigating}
& 0.350 & 0.602 & 90.78
& 0.249 & 0.727 & 88.05
& 0.438 & 0.531 & 87.92
& 0.680 & 0.225 & 84.23
& 0.304 & 0.662 & 91.65 \\
DoRA~\cite{liudora}
& 0.333 & 0.624 & 91.67
& 0.233 & 0.731 & 88.50
& 0.402 & 0.568 & 89.12
& 0.684 & 0.231 & 84.12
& 0.291 & 0.673 & 92.11 \\
\cmidrule(r){1-13} 
\cmidrule(l){14-16} 
\cellcolor{blue!10}{KDiv}
& \cellcolor{blue!10}{0.306} & \cellcolor{blue!10}{\textbf{0.667}} & \cellcolor{blue!10}{\textbf{93.75}}
& \cellcolor{blue!10}{\textbf{0.203}} & \cellcolor{blue!10}{\textbf{0.755}} & \cellcolor{blue!10}{\textbf{89.33}}
& \cellcolor{blue!10}{\textbf{0.372}} & \cellcolor{blue!10}{\textbf{0.630}} & \cellcolor{blue!10}{\textbf{89.71}}
& \cellcolor{blue!10}{\textbf{0.672}} & \cellcolor{blue!10}{0.235} & \cellcolor{blue!10}{84.40}
& \cellcolor{blue!10}{\textbf{0.268}} & \cellcolor{blue!10}{\textbf{0.693}} & \cellcolor{blue!10}{\textbf{92.97}} \\
\bottomrule[1pt]
\end{tabular}
}

\label{tab:novel}
\end{table*}

\begin{figure*}[tb]
  \centering
  \includegraphics[width=\linewidth]{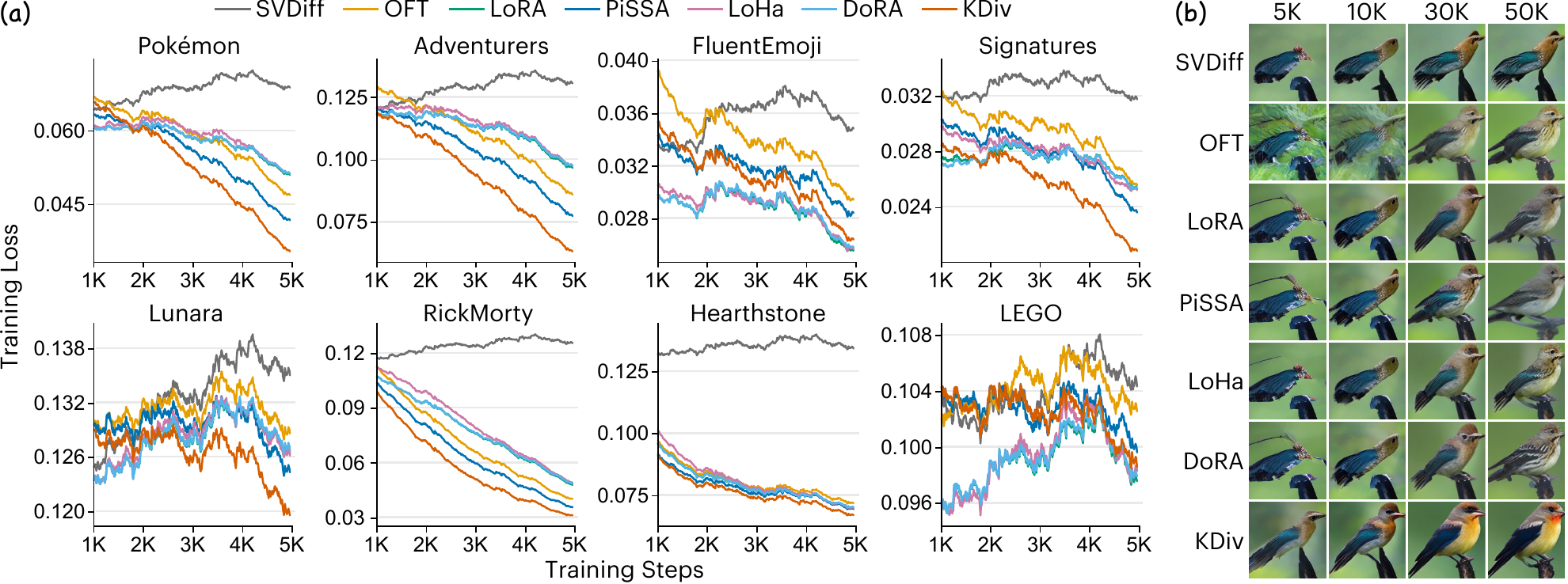}
  \vspace{-0.2in}
  \caption{
  \textbf{Downstream learning efficiency of KDiv.}
  (a)~Training loss curves across eight \textsc{Text-Conditional Generation (T2I)} datasets.
  (b)~Qualitative evolution of \textsc{Class-Conditional Generation (C2I)} samples at 5K, 10K, 30K, and 50K training steps.
  Compared with representative PEFT baselines, KDiv exhibits faster convergence and produces high-quality samples at earlier adaptation stages. Additional results are provided in Appendix Fig.~\ref{fig:app_learn}.
}
  \label{fig:fast}
\end{figure*}

\section{Results}
\subsection{Decomposable Models Pre-trained via Knowledge Diversion}
We evaluate KDiv against standard training to determine whether decomposability can be introduced without compromising generation quality. As shown in Table~\ref{tab:main_kdiv}, KDiv achieves comparable or superior performance across \textsc{Class-Conditional} (C2I), \textsc{Text-Conditional} (T2I), and \textsc{Image-Conditional} (I2I) generation, while incurring comparable or only modestly higher pre-training cost. 
Unlike dense training, which entangles condition-agnostic and condition-specific knowledge within a single parameterization, KDiv factorizes them into shared learngenes and condition-routed tailors. This design mitigates interference among heterogeneous conditional signals and promotes specialization, leading to improved distributional fidelity, compositional alignment, and conditional consistency. At inference, only the shared learngenes and selected tailors are activated, decoupling pre-training capacity from the active parameter footprint at deployment and enabling efficient condition-specific execution.

Compared with KIND~\cite{xie2025kind}, KDiv achieves superior generation quality with fewer total parameters and shorter training time. Unlike the discrete, class-specific routing adopted by KIND, KDiv employs continuous condition-aware routing to adaptively allocate knowledge across conditions. This continuity is crucial for extending knowledge diversion to \textsc{Text-Conditional} (T2I) and \textsc{Image-Conditional} (I2I) generation tasks, whose open-ended condition spaces cannot be exhaustively represented by predefined discrete classes. By avoiding rigid class-wise decomposition and the associated parameter growth, KDiv provides a more general and scalable framework for condition-aware model decomposition.

\subsection{Efficient Downstream Transfer via Adaptive Recombination}
\label{sec:downstream_transfer_results}
We investigate how the decomposition learned by KDiv facilitates transfer to novel downstream conditions.
Given a target condition, the gating network adaptively recombines the shared learngenes with relevant tailors to construct a condition-specific model.
For conditions close to the pre-training distribution, this adaptive recombination enables zero-shot generation without parameter updates (see Fig.~\ref{fig:app_i2i_zeroshot} in Appendix~\ref{sec:app_zero}). 
Under substantial distribution shifts, we freeze the learngenes and fine-tune only the target-relevant tailors, thereby preserving condition-agnostic knowledge while enabling efficient target-specific adaptation.

As shown in Table~\ref{tab:novel} and Fig.~\ref{fig:downstream}, KDiv achieves the best average performance across all reported metrics for C2I, T2I, and I2I downstream tasks, together with consistent qualitative improvements and a trainable parameter budget comparable to low-rank PEFT baselines.
Its gains over KIND on C2I further suggest that continuous routing learns more reusable components, reducing over-specialization to source classes and improving transfer to unseen targets.
More broadly, KDiv addresses a key limitation of conventional PEFT methods, which retain a frozen backbone where transferable and source-specific knowledge remain entangled. 
Their limited update parameters must therefore both suppress source-domain bias and capture target-specific variation, restricting adaptation under substantial distribution shifts. 
In contrast, KDiv preserves the condition-agnostic prior encoded by the frozen learngenes and fine-tunes only the low-rank tailors selected for the target condition. This separation concentrates optimization on target-specific knowledge, thereby mitigating negative transfer and improving both robustness and parameter efficiency.

\renewcommand{\arraystretch}{1}
\begin{table}[t]
    \centering
    \setlength{\tabcolsep}{0.52 mm}
    \caption{
    \textbf{Ablation study of knowledge diversion for downstream adaptation.} ``LG'' is the abbreviation for learngene, while ``Exclusive'' and ``Overlapping'' refer to the tailor-routing strategies adopted by KIND and KDiv, respectively. ``N/A'' denotes failure under the corresponding task setting.}
    \vspace{0.1in}
    \resizebox{\linewidth}{!}{
    \begin{tabular}{@{}lccccccc@{}}
        \toprule[1pt]
        \multicolumn{8}{c}{\textbf{\textsc{Class-Conditional Generation (C2I)}}} \\
        \midrule
        & & & \multicolumn{2}{c}{\textbf{Routing}} & \multicolumn{3}{c}{\textbf{Downstream Adaptation}} \\
        \cmidrule(lr){4-5}
        \cmidrule(lr){6-8}
        Method & LG & Tailor & Exclusive & Overlapping & FDD$\downarrow$ & KDD$\downarrow$ & CMMD$\downarrow$ \\
        \midrule
        Baseline & & & & & 1119 & 2.656 & 1.514 \\
        Only LG & \checkmark & & & & 1030 & 2.382 & 1.392 \\
        \textsc{KIND} & \checkmark & \checkmark & \checkmark & & 983 & 2.287 & 1.299 \\
        \rowcolor{blue!10}
        \textsc{KDiv} & \checkmark & \checkmark & & \checkmark & \textbf{966} & \textbf{2.239} & \textbf{1.190} \\
        \bottomrule[1pt]
        \toprule[1pt]
        \multicolumn{8}{c}{\textbf{\textsc{Text-Conditional Generation (T2I)}}} \\
        \midrule
        & & & \multicolumn{2}{c}{\textbf{Routing}} & \multicolumn{3}{c}{\textbf{Downstream Adaptation}} \\
        \cmidrule(lr){4-5}
        \cmidrule(lr){6-8}
        Method & LG & Tailor & Exclusive & Overlapping & LPIPS$\downarrow$ & SSIM$\uparrow$ & CLIP-T$\uparrow$ \\
        \midrule
        Baseline & & & & & 0.544 & 0.422 & 24.52 \\
        Only LG & \checkmark & & & & 0.539 & 0.423 & 25.08 \\
        \textsc{KIND} & \checkmark & \checkmark & \checkmark & & N/A & N/A & N/A \\
        \rowcolor{blue!10}
        \textsc{KDiv} & \checkmark & \checkmark & & \checkmark & \textbf{0.519} & \textbf{0.450} & \textbf{25.20} \\
        \bottomrule[1pt]
        \toprule[1pt]
        \multicolumn{8}{c}{\textbf{\textsc{Image-Conditional Generation (I2I)}}} \\
        \midrule
        & & & \multicolumn{2}{c}{\textbf{Routing}} & \multicolumn{3}{c}{\textbf{Downstream Adaptation}} \\
        \cmidrule(lr){4-5}
        \cmidrule(lr){6-8}
        Method & LG & Tailor & Exclusive & Overlapping & LPIPS$\downarrow$ & SSIM$\uparrow$ & CLIP-I$\uparrow$ \\
        \midrule
        Baseline & & & & & 0.292 & 0.676 & 92.09 \\
        Only LG & \checkmark & & & & 0.289 & 0.669 & 92.34 \\
        \textsc{KIND} & \checkmark & \checkmark & \checkmark & & 0.288 & 0.680 & 92.39 \\
        \rowcolor{blue!10}
        \textsc{KDiv} & \checkmark & \checkmark & & \checkmark & \textbf{0.268} & \textbf{0.693} & \textbf{92.97} \\
        \bottomrule[1pt]
    \end{tabular}
    }
    \label{tab:ablation}
\end{table}

\subsection{Ablation and Analysis}
\subsubsection{Ablation Experiments}
We assess the contributions of knowledge decomposition and tailor routing to downstream adaptation. 
As shown in Table~\ref{tab:ablation}, \textit{Only LG} outperforms the dense baseline on C2I and I2I, indicating that learngenes capture a transferable condition-agnostic prior.
Incorporating tailors with exclusive routing, as in KIND, provides additional gains by introducing condition-specific adaptation capacity. KDiv further improves performance, indicating that overlapping routing facilitates more effective sharing and composition of condition-specific knowledge than rigid exclusive assignment. 
Collectively, these results demonstrate the advantages of KDiv in knowledge diversion and continuous condition-aware routing.

\subsubsection{Strong Learning Ability Brought by Learngenes}
As shown in Fig.~\ref{fig:fast}, KDiv demonstrates superior downstream learning efficiency, achieving faster loss reduction on T2I tasks and generating higher-quality C2I samples at earlier adaptation stages. These results show that KDiv improves not only parameter efficiency but also convergence and target-specific knowledge acquisition.
In contrast, conventional PEFT methods restrict updates to a small parameter subset while retaining a largely frozen backbone in which transferable and source-specific knowledge remains entangled~\cite{ding2023parameter,han2024parameter}. KDiv instead preserves condition-agnostic knowledge in learngenes and concentrates adaptation on target-relevant tailors selected through continuous routing. This separation improves the utilization of trainable capacity, thereby accelerating convergence and enhancing downstream generation quality.

\begin{figure*}[t]
\centering
  \includegraphics[width=\linewidth]{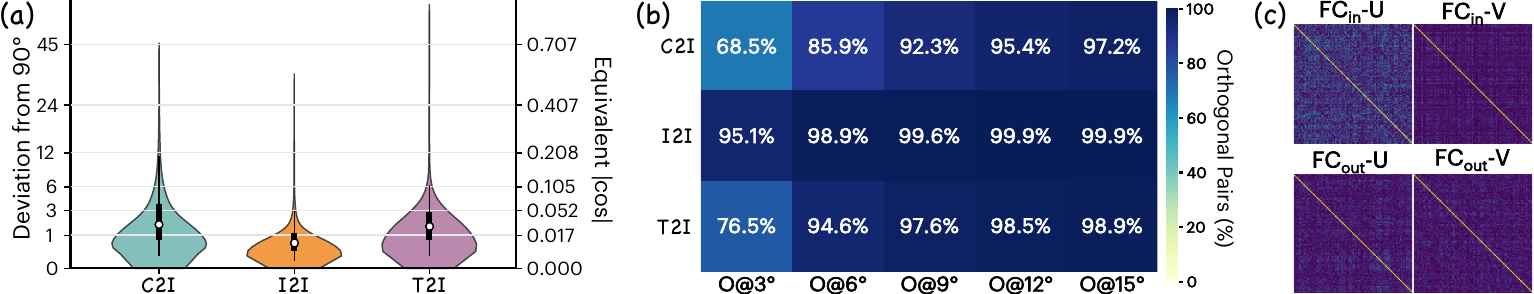}
  \vspace{-0.25in}
  \caption{
  \textbf{Emergent orthogonality among tailors.}
  (a) Distributions of pairwise angular deviations from $90^\circ$ across C2I, I2I, and T2I.
  (b) Fractions of tailor pairs satisfying different orthogonality thresholds.
  (c) Learned $U$ and $V$ matrices from representative $\mathrm{FC}_{\mathrm{in}}$ and $\mathrm{FC}_{\mathrm{out}}$ layers.
  Together, these results reveal consistently strong tailor orthogonality across conditioning modalities.
}
  \label{fig:orth}
\end{figure*}

\subsubsection{Analysis of Emergent Tailor Orthogonality}

Although SVD provides orthogonal singular-vector bases at initialization, KDiv applies no explicit orthogonality constraint during pre-training.
Nevertheless, Fig.~\ref{fig:orth}(a) shows that the pairwise angular deviations from $90^\circ$ remain concentrated near zero across C2I, T2I, and I2I. 
As quantified in Fig.~\ref{fig:orth}(b), more than $97\%$ of tailor pairs fall within $15^\circ$ of orthogonality in all three settings, with I2I preserving the initial geometry particularly well. 
Consistently, the representative $U$ and $V$ matrices in Fig.~\ref{fig:orth}(c) retain pronounced diagonal structures and limited cross-unit coupling.
This emergent orthogonality can be attributed to condition-driven routing, which exposes tailors to differentiated gradient signals and thereby mitigates redundant co-adaptation. Consequently, tailors evolve into distinct yet composable components, supporting selective routing and stable recombination without explicit orthogonality regularization.

\begin{figure*}[tb]
\centering
\includegraphics[width=\linewidth]{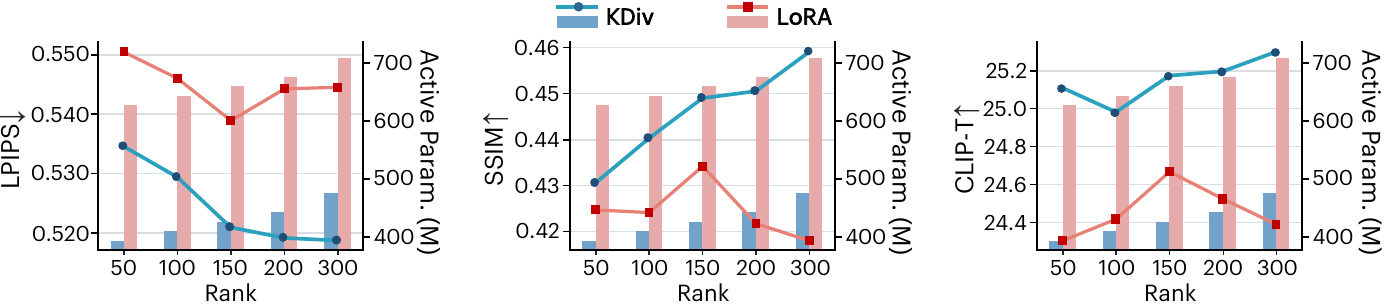}
\vspace{-0.25in}
\caption{
\textbf{Effect of tailor rank on downstream adaptation.}
Comparison of KDiv and LoRA at different ranks on T2I using LPIPS$\downarrow$, SSIM$\uparrow$, and CLIP-T$\uparrow$. Curves report generation performance, while bars indicate active parameters. KDiv consistently achieves superior performance with fewer active parameters.
}
\label{fig:rank}
\end{figure*}

\subsubsection{Analysis of Tailor Rank on Downstream Adaptation}
\label{sec:tailor_rank}
Fig.~\ref{fig:rank} examines the quality--capacity trade-off associated with tailor rank in downstream T2I adaptation. 
KDiv exhibits substantial performance gains as the tailor rank increases to 150, with only marginal improvements thereafter, indicating that a moderate rank provides sufficient capacity for target-specific adaptation.
Despite using fewer active parameters, KDiv consistently outperforms LoRA across all ranks, demonstrating that its gains stem from condition-aware knowledge reuse rather than parameter scale alone. By allocating adaptation capacity to relevant tailors, KDiv achieves a superior trade-off between generation quality and parameter efficiency.

\begin{figure}[tb]
  \centering
  \includegraphics[width=0.8\linewidth]{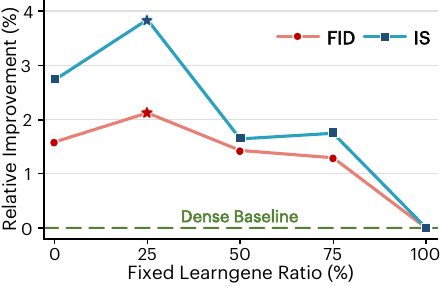}
  \vspace{-0.1in}
  \caption{\textbf{Effect of the learngene allocation on generation performance.}
  Relative improvements in FID and IS over dense training as the proportion of learngenes varies in C2I. 
  A $25\%$ learngene ratio achieves the largest gains, whereas allocating all Knowledge Units to learngenes removes condition-specific routing and recovers the dense baseline.}
  \label{fig:learngene_ratio}
\end{figure}

\subsubsection{Analysis of the Learngene Ratio}

We vary the learngene ratio while keeping the total number of Knowledge Units fixed. As shown in Fig.~\ref{fig:learngene_ratio}, performance peaks at $25\%$, demonstrating that explicitly reserving capacity for shared knowledge improves upon the tailor-only configuration.
Further increasing the ratio reduces the gains because fewer tailors remain available for condition-specific specialization. At $100\%$, all units are shared across conditions, causing KDiv to degenerate into dense training. These results underscore the importance of balancing condition-agnostic knowledge aggregation with condition-specific modeling capacity.

\subsection{Visualization of Knowledge Diversion}

\begin{figure*}[tb]
  \centering
  \includegraphics[width=\linewidth]{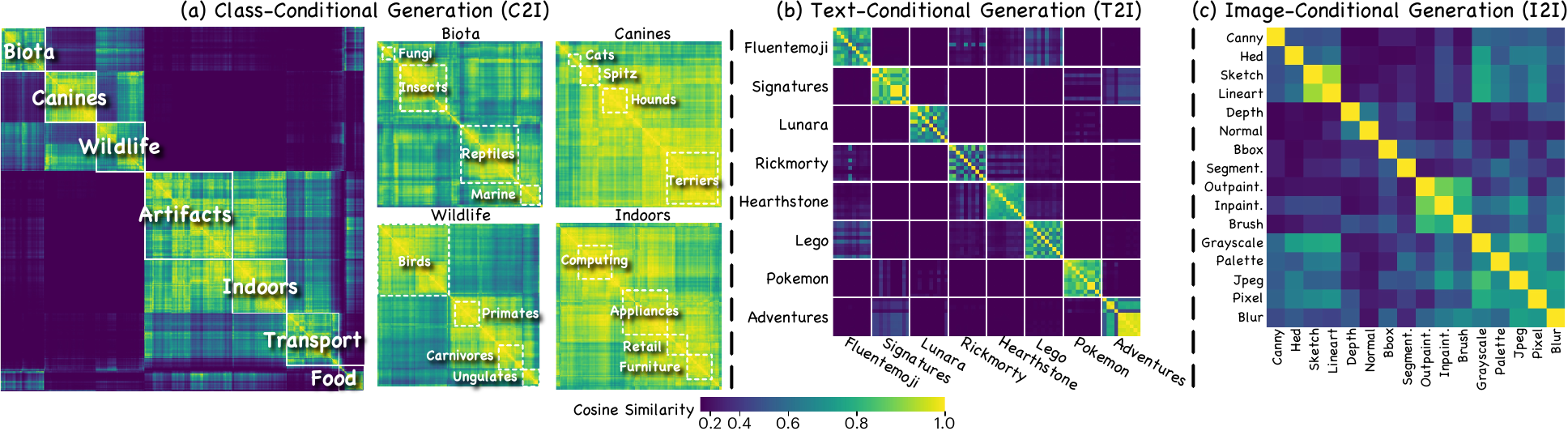}
  \vspace{-0.25in}
  \caption{
  \textbf{Condition relationships captured by the gating network.}
  Pairwise cosine similarities between the tailor modulation vectors $\boldsymbol{\sigma}_{\mathcal{T}}^{c}$ predicted by Eq.~\eqref{equ:tailor_coefficients} for conditions in (a)~C2I, (b)~T2I, and (c)~I2I. The resulting similarity structures demonstrate that the gating network captures relationships among conditions, enabling the adaptive activation and reuse of relevant tailors.
    }
  \label{fig:condition_similarity}
\end{figure*}

\begin{figure}[tb]
  \centering
  \includegraphics[width=\linewidth]{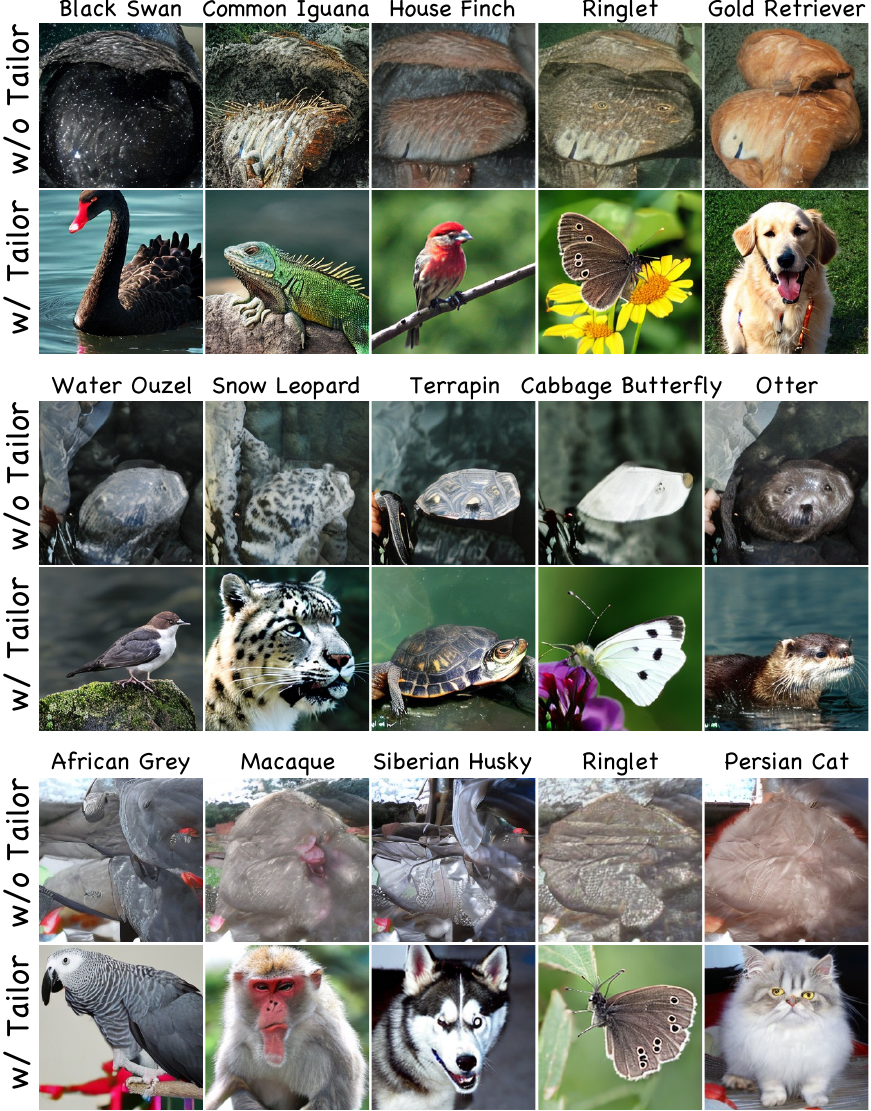}
  \vspace{-0.2in}
  \caption{Visualization of KDiv with and without tailors (i.e., learngenes only) across 15 classes and three random seeds. 
  Learngene-only generation exhibits semantic invariance, reflecting a condition-agnostic prior, whereas tailors inject condition-specific knowledge for semantically aligned generation. Additional qualitative results are provided in Appendix Fig.~\ref{fig:app_task-agnostic}.}
  \label{fig:task-agnostic}
\end{figure}

\subsubsection{Visualization of Condition-Aware Tailor Routing}
Fig.~\ref{fig:condition_similarity} visualizes the pairwise cosine similarities between the tailor modulation vectors $\boldsymbol{\sigma}_{\mathcal{T}}^{c}$. Across C2I, T2I, and I2I, semantically or structurally related conditions exhibit similar modulation patterns, yielding coherent hierarchical or block-wise structures. This organization demonstrates that the gating network captures meaningful relationships among conditions and translates them into structured tailor activation.
Instead of assigning each condition to an isolated component, KDiv enables related conditions to activate overlapping tailors. Such continuous routing encourages tailors to encode reusable condition-specific factors while maintaining distinct activation patterns for dissimilar conditions, thereby promoting knowledge sharing and facilitating adaptation to unseen conditions.

\subsubsection{Visualization of Condition-Agnostic Knowledge}
Fig.~\ref{fig:task-agnostic} provides further insight into the transferability of learngenes. In class-conditional generation, learngenes alone produce semantically invariant outputs across different class labels, indicating that they encode a condition-agnostic generative prior rather than class-specific semantics. Incorporating tailors introduces the specialized knowledge required to align generation with each target class. 
This functional separation reduces over-specialization to source conditions, enabling learngenes to serve as a broadly transferable backbone for unseen conditions.

\section{Conclusion}
In this study, we revisited image-generation pre-training from the perspective of downstream knowledge reuse and introduced Knowledge Diversion (KDiv), a condition-aware paradigm for learning decomposable generative models.
Through continuous condition-aware routing, KDiv separates condition-agnostic knowledge into learngenes and condition-specific knowledge into tailors, supporting adaptive recombination and selective downstream adaptation.
Experiments across class-, text-, and image-conditioned generation demonstrate that KDiv preserves or improves generation quality during pre-training while enabling more effective and parameter-efficient transfer to novel conditions.
Overall, these findings highlight decomposability as an effective pre-training principle for enhancing the reusability and adaptability of generative models.

\section*{Acknowledgments}
This research was supported by the Jiangsu Science Foundation (BG2024036, BK20243012), the National Natural Science Foundation of China (625B2045, 62125602, U24A20324, 92464301, 62306073), the New Cornerstone Science Foundation through the XPLORER PRIZE, the Fundamental Research Funds for the Central Universities (2242025K30024), and SEU Innovation Capability Enhancement Plan for Doctoral Students (CXJH\_SEU 26023).

\bibliographystyle{IEEEtran}
\bibliography{reference.bib}

\section{Biography Section}

\vspace{-0.35in}

\begin{IEEEbiography}[{\includegraphics[width=1in,height=1.25in,clip,keepaspectratio]{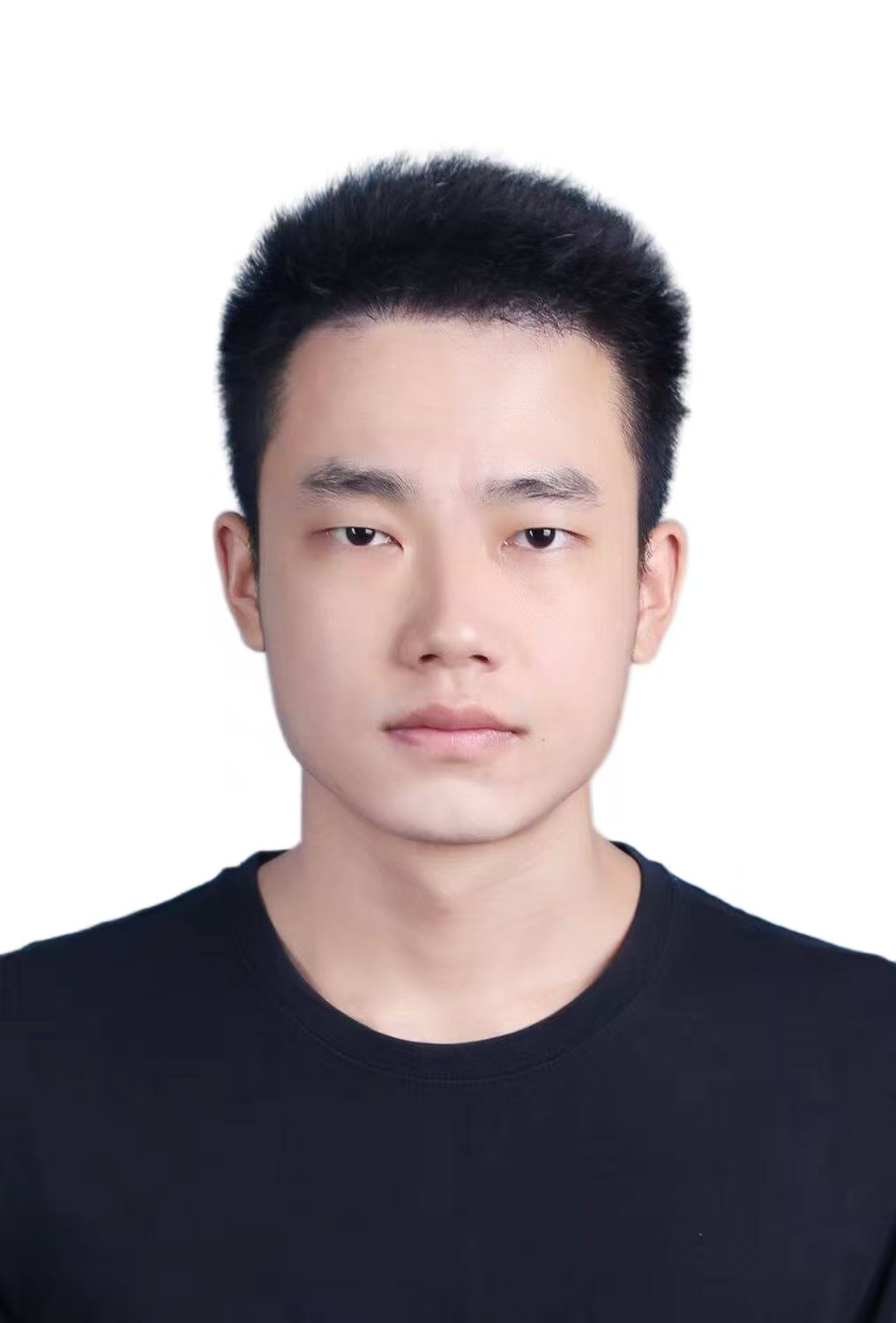}}]
{Yucheng Xie}
received the B.Sc. degree in computer science from Ocean University of China, Qingdao, China, in 2023. 
He is currently pursuing the Ph.D. degree in the School of Computer Science and Engineering, Southeast University. 
His research interests include machine learning and computer vision.
\end{IEEEbiography}

\vspace{-0.35in}

\begin{IEEEbiography}[{\includegraphics[width=1in,height=1.25in,clip,keepaspectratio]{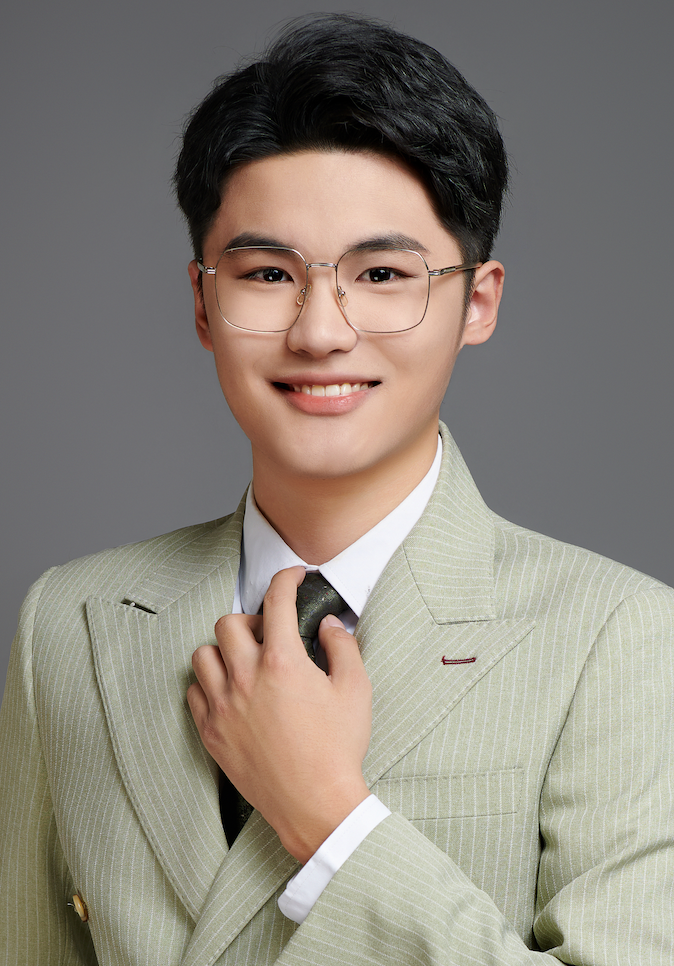}}]
{Fu Feng}
received the B.Sc. (Hons.) degree in artificial intelligence from Chien-Shiung Wu College, Southeast University, Nanjing, China, in 2023. 
He is currently pursuing the Ph.D. degree in the School of Computer Science and Engineering, Southeast University. 
His research interests include machine learning, intelligent agents and creative generation.
\end{IEEEbiography}

\newpage
\begin{IEEEbiography}[{\includegraphics[width=1in,height=1.25in,clip,keepaspectratio]{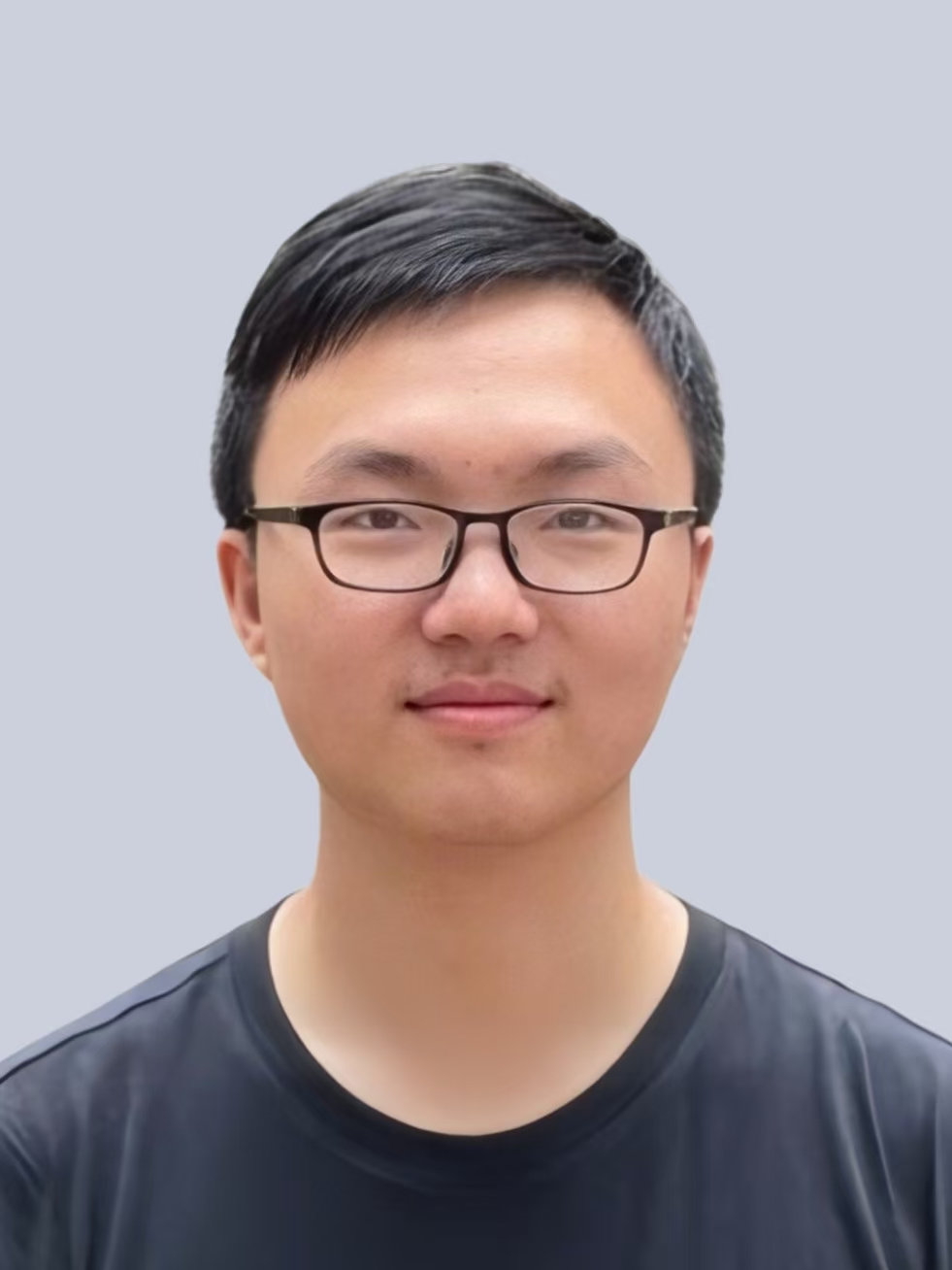}}]
{Ruixiao Shi}
received the B.Sc. degree in artificial intelligence from Southeast University, Nanjing, China, in 2025. 
He is currently pursuing the M.Sc. degree in the School of Computer Science and Engineering, Southeast University. 
His research interests include machine learning and intelligent agents.
\end{IEEEbiography}

\vspace{-0.35in}
\begin{IEEEbiography}[{\includegraphics[width=1in,height=1.25in,clip,keepaspectratio]{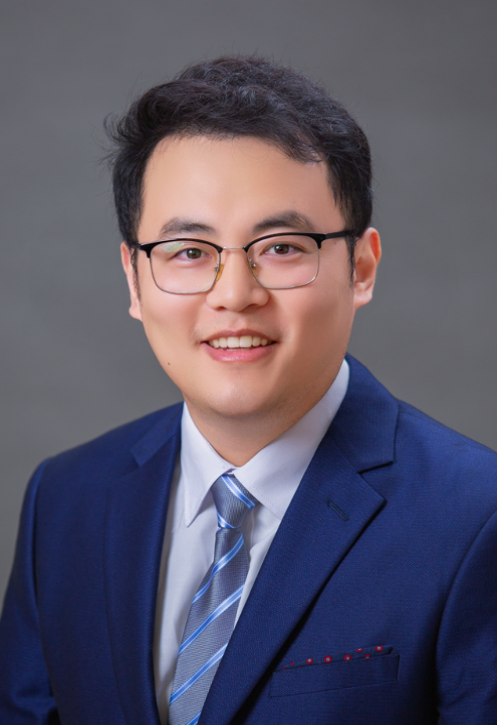}}]
{Jing Wang}
received the B.Sc. degree in computer science from Suzhou University of Science and Technology, Suzhou, China, in 2013, and the M.Sc. degree in computer science from Northeastern University, Shenyang, China, in 2015, and the Ph.D. degree in software engineering from Southeast University, Nanjing, China, in 2021. 
He is currently an assistant professor of the School of Computer Science and Engineering, Southeast University, Nanjing. His research interests include pattern recognition and machine learning.
\end{IEEEbiography}

\vspace{-0.35in}
\begin{IEEEbiography}[{\includegraphics[width=1in,height=1.25in,clip,keepaspectratio]{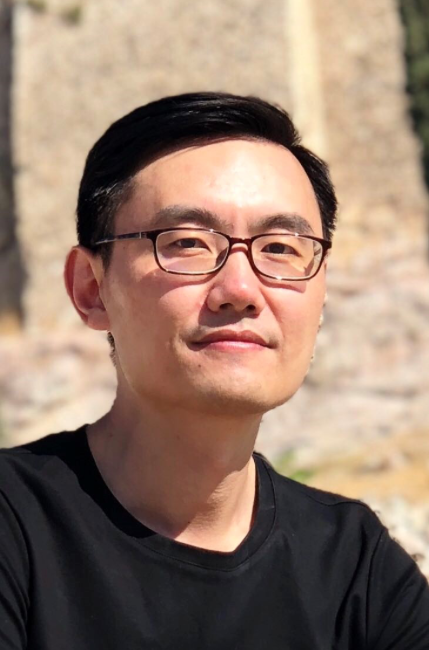}}]
{Xin Geng (Senior Member, IEEE)}
received the B.Sc. and M.Sc. degrees in computer science from Nanjing University, Nanjing, China, in 2001 and 2004, respectively, and the Ph.D. degree in computer science from Deakin University, Geelong, VIC, Australia, in 2008. 
He is currently a chair professor of the School of Computer Science and Engineering, Southeast University, Nanjing. His research interests include machine learning, pattern recognition, and computer vision. He has published over 100 refereed articles in these areas, including those published in prestigious journals and top international conferences. 

Dr. Geng has been an Associate Editor of IEEE TRANSACTIONS ON MULTIMEDIA, Frontiers of Computer Science, and Mathematical Foundations of Computing, a Steering Committee Member of Pacific Rim International Conferences on Artificial Intelligence (PRICAI), a Program Committee Chair for conferences, such as PRICAI 2018 and  Vision And Learning SEminar (VALSE) 2013, the Area Chair for conferences, such as Computer Vision and Pattern Recognition (CVPR), ACM Multimedia, and Chinese Conference on Pattern Recognition (CCPR), and a Senior Program Committee Member for conferences, such as International Joint Conference on Artificial Intelligence (IJCAI), AAAI Conference on Artificial Intelligence (AAAI), and European Conference on Artificial Intelligence (ECAI). He is a Distinguished Fellow of International Engineering and Technology Institute. 
\end{IEEEbiography}

\vspace{-0.35in}
\begin{IEEEbiography}[{\includegraphics[width=1in,height=1.25in,clip,keepaspectratio]{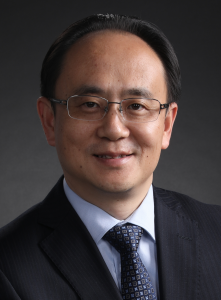}}]
{Yong Rui (Fellow, IEEE)} 
received the BS degree from Southeast University, the MS degree from Tsinghua University, and the PhD degree from the University of Illinois at Urbana-Champaign. He is currently the corporate chief technology officer and the senior vice president with Lenovo Group. 
He holds 112 issued patents, has published 4 books, 12 book chapters, and 300 journal and conference papers. With more than 40,550 citations, and an h-Index of 94, his publications are among the most referenced. 
He holds 70 issued U.S. and international patents. He, a noted researcher, was a recipient of the 2018 ACM SIGMM Technical Achievement Award, the 2016 IEEE Computer Society Edward J. McCluskey Technical Achievement Award, the 2016 IEEE Signal Processing Society Best Paper Award, and the 2010 Most Cited Paper of the Decade Award from Journal of Visual Communication and Image Representation. He is an associate editor of ACM Transactions on Multimedia Computing, Communication and Applications (TOMM), and a founding editor of International Journal of Multimedia Information Retrieval. 
He was the editor-in-chief of the IEEE MultiMedia Magazine (2013–2017), an associate editor of the IEEE Transactions On Multimedia (2004–2008), the IEEE Transactions on Circuits and Systems for Video Technologies(2006–2010), ACM/Springer Multimedia Systems Journal(2004–2006), International Journal of Multimedia Tools and Applications (2004–2006), and the IEEE Access (2013–2016). 
He is a fellow of ACM, IAPR, and SPIE, and a foreign member of Academia Europaea and Canadian Academy of Engineering.
\end{IEEEbiography}

\clearpage
\appendices
\onecolumn

\section{Additional Related Work}
\label{sec:app_related}
\subsection{Conditional Image Generation: Tasks and Models}

Conditional image generation synthesizes visual content under semantic or structural guidance, with tasks commonly distinguished by the conditioning modality. \textsc{Class-Conditional Generation} uses discrete labels to specify semantic categories, providing a controlled setting for evaluating generation quality and model scalability~\citesupp{supp_dhariwal2021diffusion,supp_peebles2023scalable}. \textsc{Text-Conditional Generation} translates open-ended natural-language descriptions into images, enabling substantially richer semantic control~\citesupp{supp_nichol2022glide,supp_ramesh2022hierarchical,supp_saharia2022photorealistic}. \textsc{Image-Conditional Generation} instead uses reference images or spatial signals, such as edges, poses, depth maps, and segmentation masks, to preserve visual structure during translation, editing, and controllable synthesis~\citesupp{supp_meng2021sdedit,supp_zhang2023adding,supp_mou2024t2i}. These settings differ not only in input representation but also in the knowledge required to interpret and satisfy each condition, motivating a unified yet condition-aware generative framework.

High-fidelity image synthesis was initially driven by generative adversarial networks~\citesupp{supp_brock2018large,supp_karras2019style}. Diffusion models subsequently became a prevailing alternative because of their strong generation quality, stable optimization, and flexibility in incorporating diverse conditions~\citesupp{supp_song2021scorebased,supp_ho2021classifierfree}. Early diffusion models primarily employed convolutional U-Nets for iterative denoising~\citesupp{supp_ho2020denoising,supp_dhariwal2021diffusion}. Latent diffusion models reduced the computational burden by moving the generative process from pixel space to a compressed representation space~\citesupp{supp_rombach2022high}, while Transformer-based architectures further improved scalability and facilitated unified conditioning~\citesupp{supp_bao2023all,supp_peebles2023scalable}.
Recent studies have advanced diffusion modeling through unified diffusion--flow formulations~\citesupp{supp_ma2024sit}, alternative prediction objectives~\citesupp{supp_li2025back}, alignment with pretrained visual representations~\citesupp{supp_yu2024representation}, and efficient one-step generation~\citesupp{supp_geng2026mean}. These developments primarily improve model architecture, training objectives, or sampling efficiency while retaining a unified parameterization in which knowledge associated with different conditions remains entangled. KDiv is complementary to these directions: it focuses on how conditional knowledge is organized during pre-training, enabling the resulting model to be decomposed and recombined according to downstream requirements.

\section{Theoretical Insights into Knowledge Diversion}
\label{sec:theoretical_insights}

We characterize the KDiv parameterization from three complementary perspectives. First, we compare continuous tailor routing with exclusive assignment and show that the latter constitutes a restricted case of the former under the same set of Knowledge Units. Second, we analyze the approximation error introduced by top-$K$ tailor selection, providing a structural justification for modulation-based model recombination. Third, we examine the continuity of condition-aware recombination and establish how variations in condition representations affect the resulting model parameters. Together, these analyses clarify the representational flexibility, compression behavior, and stability of KDiv.

We analyze an arbitrary projection matrix $W_{\star}^{(l)}$, whose effective weight under condition $c$ is reconstructed as
\begin{equation}
\widetilde{W}_{\star,c}^{(l)}
=
\mathcal{G}_{\star}^{(l)}
+
\sum_{i=1}^{N_T}
\left[\boldsymbol{\sigma}_{\mathcal{T}}^{c}\right]_i
\mathcal{T}_{\star,i}^{(l)},
\label{equ:theory_reconstruction}
\end{equation}
where $\mathcal{G}_{\star}^{(l)}$ is the learngene of $W_{\star}^{(l)}$, obtained by aggregating its $N_G$ Knowledge Units:
\begin{equation*}
\mathcal{G}_{\star}^{(l)}
=
\sum_{j=1}^{N_G}
\mathbf{u}_{\star}^{(l,j)}
{\mathbf{v}_{\star}^{(l,j)}}^{\top}.
\end{equation*}
Similarly, $\mathcal{T}_{\star,i}^{(l)}$ denotes the $i$-th Knowledge Unit of its tailor:
\begin{equation*}
\mathcal{T}_{\star,i}^{(l)}
=
\mathbf{u}_{\star}^{(l,N_G+i)}
{\mathbf{v}_{\star}^{(l,N_G+i)}}^{\top}.
\end{equation*}
The condition-dependent modulation vector
$\boldsymbol{\sigma}_{\mathcal{T}}^{c}
=
\mathcal{H}(\mathcal{E}_{\mathrm{cond}}(c))$
determines the contribution of each tailor Knowledge Unit to the reconstructed weight. The following analysis examines the structural properties induced by this parameterization.

\subsection{Expressiveness of Continuous Tailor Routing}
\label{sec:theory_expressiveness}

We first compare the representational capacity of continuous overlapping routing with that of exclusive routing. Let
$\mathbf{e}_i\in\mathbb{R}^{N_T}$ denote the $i$-th standard basis vector. An exclusive routing mechanism activates at most one tailor for each condition and can therefore be represented by modulation vectors in
\begin{equation*}
\mathcal{A}_{\mathrm{exc}}
=
\left\{
\alpha\mathbf{e}_i
\;\middle|\;
\alpha\in\mathbb{R},\;
i\in[1,N_T]
\right\}.
\end{equation*}
This definition allows the selected component to have an arbitrary scalar modulation $\alpha$ and is therefore at least as general as a binary one-hot gate. In contrast, continuous overlapping routing permits modulation vectors in
\begin{equation*}
\mathcal{A}_{\mathrm{cont}}
=
\mathbb{R}^{N_T},
\end{equation*}
allowing multiple tailors to contribute simultaneously. At the level of routing parameterization,
\begin{equation*}
\mathcal{A}_{\mathrm{exc}}
\subseteq
\mathcal{A}_{\mathrm{cont}}.
\end{equation*}

To make this relationship explicit, vectorizing Eq.~\eqref{equ:theory_reconstruction} gives
\begin{equation}
\operatorname{vec}\!\left(
\widetilde{W}_{\star,c}^{(l)}
\right)
=
\operatorname{vec}\!\left(
\mathcal{G}_{\star}^{(l)}
\right)
+
B_{\star}^{(l)}
\boldsymbol{\sigma}_{\mathcal{T}}^{c},
\label{equ:theory_vectorized}
\end{equation}
where
\begin{equation*}
B_{\star}^{(l)}
=
\left[
\operatorname{vec}\!\left(\mathcal{T}_{\star,1}^{(l)}\right),
\ldots,
\operatorname{vec}\!\left(\mathcal{T}_{\star,N_T}^{(l)}\right)
\right].
\end{equation*}
Under exclusive routing, the reconstructed weight is restricted to the union of the one-dimensional affine sets
\begin{equation*}
\operatorname{vec}\!\left(
\mathcal{G}_{\star}^{(l)}
\right)
+
\operatorname{span}
\left\{
\operatorname{vec}\!\left(
\mathcal{T}_{\star,i}^{(l)}
\right)
\right\}.
\end{equation*}
Continuous routing instead allows reconstruction within the affine subspace
\begin{equation*}
\operatorname{vec}\!\left(
\mathcal{G}_{\star}^{(l)}
\right)
+
\operatorname{span}
\left\{
\operatorname{vec}\!\left(
\mathcal{T}_{\star,i}^{(l)}
\right)
\right\}_{i=1}^{N_T}.
\end{equation*}
Its effective dimension is
$\operatorname{rank}(B_{\star}^{(l)})\leq N_T$.

Therefore, exclusive routing is a special case of the continuous formulation, provided that the gating network can represent the corresponding one-sparse modulation vectors. The inclusion becomes strict when at least two tailors are linearly independent and the gating network can assign them nonzero values simultaneously. In this case, their joint contribution cannot generally be represented by selecting either tailor alone.

This result establishes a representational advantage rather than a performance guarantee. Continuous routing provides a larger space of condition-specific weight compositions, but the usefulness of this additional flexibility still depends on how effectively the gating network and Knowledge Units are optimized. Nevertheless, it explains why KDiv can reuse overlapping condition-specific factors, whereas exclusive routing imposes a rigid assignment between conditions and individual components.

\subsection{Approximation Error of Tailor Selection}
\label{sec:theory_selection_error}

For downstream adaptation, KDiv retains only the $K$ tailors most relevant to the target condition distribution. Let
$\mathcal{I}_K\subseteq[1,N_T]$ denote the selected index set, with
$|\mathcal{I}_K|=K$. The reconstructed weight after selection is
\begin{equation*}
\widetilde{W}_{\star,c}^{(l,K)}
=
\mathcal{G}_{\star}^{(l)}
+
\sum_{i\in\mathcal{I}_K}
\left[\boldsymbol{\sigma}_{\mathcal{T}}^{c}\right]_i
\mathcal{T}_{\star,i}^{(l)}.
\end{equation*}
The difference from the complete reconstruction is
\begin{equation*}
R_{\star,c}^{(l,K)}
=
\widetilde{W}_{\star,c}^{(l)}
-
\widetilde{W}_{\star,c}^{(l,K)}
=
\sum_{i\notin\mathcal{I}_K}
\left[\boldsymbol{\sigma}_{\mathcal{T}}^{c}\right]_i
\mathcal{T}_{\star,i}^{(l)}.
\end{equation*}

Without assuming orthogonality among tailors, the triangle inequality yields
\begin{equation}
\left\|
R_{\star,c}^{(l,K)}
\right\|_{\mathrm{F}}
\leq
\sum_{i\notin\mathcal{I}_K}
\left|
\left[\boldsymbol{\sigma}_{\mathcal{T}}^{c}\right]_i
\right|
\left\|
\mathcal{T}_{\star,i}^{(l)}
\right\|_{\mathrm{F}}.
\label{equ:theory_general_error_bound}
\end{equation}
This bound shows that removing a tailor has a limited effect when either its modulation magnitude or its matrix norm is small.

For a target condition set $\mathcal{C}_{\mathrm{tgt}}$, recall that KDiv estimates the relevance of the $i$-th tailor by
\begin{equation*}
s_i(\mathcal{C}_{\mathrm{tgt}})
=
\frac{1}{|\mathcal{C}_{\mathrm{tgt}}|}
\sum_{c\in\mathcal{C}_{\mathrm{tgt}}}
\left|
\left[\boldsymbol{\sigma}_{\mathcal{T}}^{c}\right]_i
\right|.
\end{equation*}
Averaging Eq.~\eqref{equ:theory_general_error_bound} over the target conditions gives
\begin{equation}
\frac{1}{|\mathcal{C}_{\mathrm{tgt}}|}
\sum_{c\in\mathcal{C}_{\mathrm{tgt}}}
\left\|
R_{\star,c}^{(l,K)}
\right\|_{\mathrm{F}}
\leq
\sum_{i\notin\mathcal{I}_K}
s_i(\mathcal{C}_{\mathrm{tgt}})
\left\|
\mathcal{T}_{\star,i}^{(l)}
\right\|_{\mathrm{F}}.
\label{equ:theory_expected_error_bound}
\end{equation}
If the tailor norms are equal or remain approximately comparable, selecting the $K$ largest relevance scores minimizes the right-hand side among all subsets of size $K$. More generally, when the norms vary substantially, the corresponding bound is minimized by ranking tailors according to
$s_i(\mathcal{C}_{\mathrm{tgt}})
\|\mathcal{T}_{\star,i}^{(l)}\|_{\mathrm{F}}$.
Thus, the modulation-based criterion used by KDiv can be interpreted as a norm-independent approximation to this weighted selection rule. This interpretation is particularly appropriate when the factor units remain similarly scaled.

A stronger result can be obtained under orthogonality. Suppose that the tailor components associated with a given weight matrix are orthonormal under the Frobenius inner product:
\begin{equation*}
\left\langle
\mathcal{T}_{\star,i}^{(l)},
\mathcal{T}_{\star,j}^{(l)}
\right\rangle_{\mathrm{F}}
=
\mathbb{I}[i=j].
\end{equation*}
Then the squared reconstruction error becomes
\begin{equation*}
\left\|
R_{\star,c}^{(l,K)}
\right\|_{\mathrm{F}}^{2}
=
\sum_{i\notin\mathcal{I}_K}
\left|
\left[\boldsymbol{\sigma}_{\mathcal{T}}^{c}\right]_i
\right|^{2}.
\end{equation*}
For an individual condition, retaining the $K$ largest modulation magnitudes therefore minimizes the squared reconstruction error exactly. Over a target condition set, the corresponding expected squared error is minimized by ranking tailors according to their root-mean-square modulation values. The mean-magnitude score adopted by KDiv instead minimizes the general upper bound in Eq.~\eqref{equ:theory_expected_error_bound} when the unit norms are comparable.

Exact orthogonality is not enforced during KDiv training. To characterize the approximate case, define the mutual coherence
\begin{equation*}
\mu_{\star}^{(l)}
=
\max_{i\neq j}
\frac{
\left|
\left\langle
\mathcal{T}_{\star,i}^{(l)},
\mathcal{T}_{\star,j}^{(l)}
\right\rangle_{\mathrm{F}}
\right|
}{
\left\|
\mathcal{T}_{\star,i}^{(l)}
\right\|_{\mathrm{F}}
\left\|
\mathcal{T}_{\star,j}^{(l)}
\right\|_{\mathrm{F}}
}.
\end{equation*}
Writing
$a_i=[\boldsymbol{\sigma}_{\mathcal{T}}^{c}]_i$,
the deviation from the orthogonal error decomposition satisfies
\begin{equation*}
\left|
\left\|
R_{\star,c}^{(l,K)}
\right\|_{\mathrm{F}}^{2}
-
\sum_{i\notin\mathcal{I}_K}
|a_i|^{2}
\left\|
\mathcal{T}_{\star,i}^{(l)}
\right\|_{\mathrm{F}}^{2}
\right|
\quad\leq
\mu_{\star}^{(l)}
\sum_{\substack{i,j\notin\mathcal{I}_K\\i\neq j}}
|a_i a_j|
\left\|
\mathcal{T}_{\star,i}^{(l)}
\right\|_{\mathrm{F}}
\left\|
\mathcal{T}_{\star,j}^{(l)}
\right\|_{\mathrm{F}}.
\end{equation*}
Hence, low coherence reduces the contribution of cross-unit interactions and makes modulation magnitude a more reliable indicator of reconstruction importance. This provides a structural connection between the emergent tailor orthogonality observed in Fig.~\ref{fig:orth} and the effectiveness of top-$K$ model recombination.

These bounds concern approximation in parameter space and do not imply that minimizing reconstruction error necessarily minimizes the downstream diffusion objective. They nevertheless provide a direct justification for using the learned modulation magnitudes as relevance scores when constructing a compact target model.

\subsection{Continuity of Condition-Aware Recombination}
\label{sec:theory_continuity}

We next examine how changes in the condition representation affect the reconstructed model parameters. Consider two conditions $c$ and $c'$. Because the learngene component is shared and remains unchanged, subtracting their reconstructed weights gives
\begin{equation*}
\widetilde{W}_{\star,c}^{(l)}
-
\widetilde{W}_{\star,c'}^{(l)}
=
\sum_{i=1}^{N_T}
\left(
\left[\boldsymbol{\sigma}_{\mathcal{T}}^{c}\right]_i
-
\left[\boldsymbol{\sigma}_{\mathcal{T}}^{c'}\right]_i
\right)
\mathcal{T}_{\star,i}^{(l)}.
\end{equation*}
Applying the triangle and Cauchy--Schwarz inequalities yields
\begin{equation}
\left\|
\widetilde{W}_{\star,c}^{(l)}
-
\widetilde{W}_{\star,c'}^{(l)}
\right\|_{\mathrm{F}}
\leq
\sum_{i=1}^{N_T}
\left|
\left[\boldsymbol{\sigma}_{\mathcal{T}}^{c}\right]_i
-
\left[\boldsymbol{\sigma}_{\mathcal{T}}^{c'}\right]_i
\right|
\left\|
\mathcal{T}_{\star,i}^{(l)}
\right\|_{\mathrm{F}}
\leq
\left\|
\boldsymbol{\sigma}_{\mathcal{T}}^{c}
-
\boldsymbol{\sigma}_{\mathcal{T}}^{c'}
\right\|_2
\left(
\sum_{i=1}^{N_T}
\left\|
\mathcal{T}_{\star,i}^{(l)}
\right\|_{\mathrm{F}}^2
\right)^{1/2}.
\label{equ:theory_continuity_bound}
\end{equation}
Therefore, conditions with similar tailor modulation vectors necessarily induce nearby effective weights, with the sensitivity controlled by the aggregate norm of the tailor components.

This result can also be expressed over the complete collection of reconstructed matrices. Define
\begin{equation*}
\left\|
\widetilde{\Theta}_{c}
-
\widetilde{\Theta}_{c'}
\right\|_{\mathrm{F}}^{2}
=
\sum_{l=1}^{L}
\sum_{\star\in\mathcal{S}}
\left\|
\widetilde{W}_{\star,c}^{(l)}
-
\widetilde{W}_{\star,c'}^{(l)}
\right\|_{\mathrm{F}}^{2}
\end{equation*}
and the total tailor magnitude
\begin{equation*}
B_{\mathcal{T}}
=
\left(
\sum_{l=1}^{L}
\sum_{\star\in\mathcal{S}}
\sum_{i=1}^{N_T}
\left\|
\mathcal{T}_{\star,i}^{(l)}
\right\|_{\mathrm{F}}^{2}
\right)^{1/2}.
\end{equation*}
Eq.~\eqref{equ:theory_continuity_bound} then implies
\begin{equation}
\left\|
\widetilde{\Theta}_{c}
-
\widetilde{\Theta}_{c'}
\right\|_{\mathrm{F}}
\leq
B_{\mathcal{T}}
\left\|
\boldsymbol{\sigma}_{\mathcal{T}}^{c}
-
\boldsymbol{\sigma}_{\mathcal{T}}^{c'}
\right\|_2.
\label{equ:theory_model_continuity}
\end{equation}

Let $\mathbf{e}_c=\mathcal{E}_{\mathrm{cond}}(c)$ denote the representation produced by the corresponding modality-specific encoder. If the gating network $\mathcal{H}$ is $L_{\mathcal{H}}$-Lipschitz within this representation space, then
\begin{equation*}
\left\|
\boldsymbol{\sigma}_{\mathcal{T}}^{c}
-
\boldsymbol{\sigma}_{\mathcal{T}}^{c'}
\right\|_2
\leq
L_{\mathcal{H}}
\left\|
\mathbf{e}_c-\mathbf{e}_{c'}
\right\|_2.
\end{equation*}
Combining this relation with Eq.~\eqref{equ:theory_model_continuity} gives
\begin{equation}
\left\|
\widetilde{\Theta}_{c}
-
\widetilde{\Theta}_{c'}
\right\|_{\mathrm{F}}
\leq
B_{\mathcal{T}}L_{\mathcal{H}}
\left\|
\mathbf{e}_c-\mathbf{e}_{c'}
\right\|_2.
\label{equ:theory_lipschitz_recombination}
\end{equation}
Under this assumption, nearby condition representations induce bounded changes in the recombined model parameters. The bound applies within each modality-specific representation space and does not require representations from different modalities to share a common embedding space.

This continuity provides a theoretical interpretation of the structured modulation similarities observed in Fig.~\ref{fig:condition_similarity}. When related conditions are mapped to nearby modulation vectors, they activate similar combinations of tailors and consequently share condition-specific knowledge. Unlike one-hot routing, continuous routing permits gradual variation in the effective model as the condition representation changes.

The analysis establishes continuity of the parameter mapping, rather than directly guaranteeing similarity between generated images. Such a guarantee would additionally require assumptions on the stability of the denoising network and the sampling process. Moreover, downstream top-$K$ selection introduces a discrete mask. The continuity result therefore holds globally for the soft-routing model and locally for a recombined target model while its selected tailor set remains fixed. Since KDiv determines this set once from the target condition distribution, continuous modulation is preserved among the retained tailors during downstream adaptation.

Together, these results characterize three complementary properties of KDiv. Continuous routing provides a richer composition space than exclusive assignment, modulation-based selection controls the parameter reconstruction error under identifiable conditions, and similar modulation patterns induce nearby condition-specific parameterizations. These properties offer a structural explanation for the empirical flexibility of KDiv without requiring stronger claims about optimization or generalization.

\section{Additional Methodological Details}
\subsection{Gradient Interpretation of Knowledge Diversion}
\label{sec:app_gradient}

We examine the optimization dynamics of KDiv to explain how condition-aware routing separates condition-agnostic knowledge from condition-specific information. For an arbitrary projection matrix $W_{\star}^{(l)}$, the condition-specific weight used in forward propagation is written as
\begin{equation*}
\widetilde{W}_{\star,c}^{(l)}
=
\mathcal{G}_{\star}^{(l)}
+
\sum_{i=1}^{N_T}
\left[\boldsymbol{\sigma}_{\mathcal{T}}^{c}\right]_i
\mathcal{T}_{\star,i}^{(l)},
\end{equation*}
where the aggregated learngene component is
\begin{equation*}
\mathcal{G}_{\star}^{(l)}
=
\sum_{j=1}^{N_G}
\mathbf{u}_{\star}^{(l,j)}
{\mathbf{v}_{\star}^{(l,j)}}^{\top},
\end{equation*}
and the $i$-th tailor component is
\begin{equation*}
\mathcal{T}_{\star,i}^{(l)}
=
\mathbf{u}_{\star}^{(l,N_G+i)}
{\mathbf{v}_{\star}^{(l,N_G+i)}}^{\top},
\qquad i=1,\ldots,N_T.
\end{equation*}
Thus, $\mathcal{G}_{\star}^{(l)}$ aggregates the $N_G$ Knowledge Units in $\mathcal{G}$ associated with $W_{\star}^{(l)}$, whereas $\mathcal{T}_{\star,i}^{(l)}$ denotes its $i$-th Knowledge Unit drawn from $\mathcal{T}$. The learngene component remains active for every condition, while each tailor contribution is determined by the corresponding modulation value $\left[\boldsymbol{\sigma}_{\mathcal{T}}^{c}\right]_i$ predicted by the gating network.

Let
\begin{equation*}
\Delta_{\star,c}^{(l)}
=
\frac{\partial\mathcal{L}_{\mathrm{diff}}}
{\partial\widetilde{W}_{\star,c}^{(l)}}
\end{equation*}
denote the gradient propagated to the reconstructed weight matrix. Applying the chain rule, the gradients of the $j$-th learngene factors are
\begin{equation*}
\frac{\partial\mathcal{L}_{\mathrm{diff}}}
{\partial\mathbf{u}_{\star}^{(l,j)}}
=
\Delta_{\star,c}^{(l)}
\mathbf{v}_{\star}^{(l,j)},
\qquad
\frac{\partial\mathcal{L}_{\mathrm{diff}}}
{\partial\mathbf{v}_{\star}^{(l,j)}}
=
{\Delta_{\star,c}^{(l)}}^{\top}
\mathbf{u}_{\star}^{(l,j)}.
\end{equation*}
Because these gradients are independent of the condition-dependent modulation values, every training sample contributes directly to the optimization of the learngenes. Their expected updates therefore aggregate learning signals across the complete condition distribution, encouraging $\mathcal{G}$ to capture generative knowledge shared across conditions rather than information specific to individual conditions.

In contrast, the optimization of each tailor is explicitly conditioned on its predicted modulation value. For the $i$-th tailor, corresponding to the original factor index $N_G+i$, the gradients of its left and right factor vectors are
\begin{equation*}
\frac{\partial\mathcal{L}_{\mathrm{diff}}}
{\partial\mathbf{u}_{\star}^{(l,N_G+i)}}
=
\left[\boldsymbol{\sigma}_{\mathcal{T}}^{c}\right]_i
\Delta_{\star,c}^{(l)}
\mathbf{v}_{\star}^{(l,N_G+i)},
\qquad
\frac{\partial\mathcal{L}_{\mathrm{diff}}}
{\partial\mathbf{v}_{\star}^{(l,N_G+i)}}
=
\left[\boldsymbol{\sigma}_{\mathcal{T}}^{c}\right]_i
{\Delta_{\star,c}^{(l)}}^{\top}
\mathbf{u}_{\star}^{(l,N_G+i)}.
\end{equation*}
Unlike learngenes, whose gradients are contributed by every condition without modulation, each tailor receives a condition-dependent update scaled by $\left[\boldsymbol{\sigma}_{\mathcal{T}}^{c}\right]_i$. A larger modulation magnitude amplifies the gradient propagated to the corresponding tailor, whereas a value close to zero suppresses its update. Consequently, each condition primarily optimizes the tailors most relevant to its generative characteristics.

The modulation values are learned jointly with the Knowledge Units. Because each entry of $\boldsymbol{\sigma}_{\mathcal{T}}^{c}$ controls the corresponding tailor across the factorized weight matrices, its gradient accumulates the alignment between that tailor and the propagated gradients:
\begin{equation*}
\frac{\partial\mathcal{L}_{\mathrm{diff}}}
{\partial
\left[\boldsymbol{\sigma}_{\mathcal{T}}^{c}\right]_i}
=
\sum_{l=1}^{L}
\sum_{\star\in\mathcal{S}}
\left\langle
\Delta_{\star,c}^{(l)},
\mathcal{T}_{\star,i}^{(l)}
\right\rangle_{\mathrm{F}},
\end{equation*}
where $\langle\cdot,\cdot\rangle_{\mathrm{F}}$ denotes the Frobenius inner product. This gradient measures how strongly the $i$-th tailor aligns with the update required by condition $c$. Tailors that better support the current condition therefore receive modulation values that exert greater influence on the reconstructed weights.

Let $\boldsymbol{\phi}_{\mathcal{H}}$ denote the parameters of the gating network $\mathcal{H}$. Since
$\boldsymbol{\sigma}_{\mathcal{T}}^{c}
=
\mathcal{H}(\mathbf{e}_c)$,
the gradient propagated to $\mathcal{H}$ is
\begin{equation*}
\frac{\partial\mathcal{L}_{\mathrm{diff}}}
{\partial\boldsymbol{\phi}_{\mathcal{H}}}
=
\sum_{i=1}^{N_T}
\frac{\partial\mathcal{L}_{\mathrm{diff}}}
{\partial
\left[\boldsymbol{\sigma}_{\mathcal{T}}^{c}\right]_i}
\frac{\partial
\left[\boldsymbol{\sigma}_{\mathcal{T}}^{c}\right]_i}
{\partial\boldsymbol{\phi}_{\mathcal{H}}}.
\end{equation*}
The gating network is thus optimized to assign each condition an appropriate pattern of tailor contributions rather than an exclusive component assignment.

These asymmetric gradient pathways provide the optimization basis for knowledge diversion. Learngenes aggregate unmodulated gradients across the condition distribution and consequently capture broadly reusable generative knowledge. Tailors, by contrast, receive selectively weighted gradients and specialize in condition-dependent variations. Conditions that induce similar modulation patterns jointly update overlapping tailors, allowing related conditions to reuse specialized knowledge while retaining distinct combinations for their individual requirements.

\setlength{\algomargin}{0pt}
\begin{algorithm}[t]
\SetAlCapHSkip{0pt}
\SetInd{0.4em}{0.6em}
\caption{Target-Aware Model Recombination}
\label{alg:target_recombination}

\KwIn{Target dataset $\mathcal{D}_{\mathrm{tgt}}$;
pretrained learngenes $\mathcal{G}$, tailors $\mathcal{T}$,
condition encoder $\mathcal{E}_{\mathrm{cond}}$, and gating network
$\mathcal{H}$; number of selected tailors $K$}

\KwOut{Recombined target model $\mathcal{M}_{\mathrm{tgt}}$
with parameters $\Theta_{\mathrm{tgt}}$}

\BlankLine
\textbf{Target-Aware Tailor Selection}\;

Extract the target condition set
$\mathcal{C}_{\mathrm{tgt}}$ from $\mathcal{D}_{\mathrm{tgt}}$\;

\ForEach{$c\in\mathcal{C}_{\mathrm{tgt}}$}{
    Compute the condition representation
    $\mathbf{e}_c
    \gets
    \mathcal{E}_{\mathrm{cond}}(c)$\;
    Predict the tailor modulation values
    $\boldsymbol{\sigma}_{\mathcal{T}}^{c}
    \gets
    \mathcal{H}(\mathbf{e}_c)$\;
}

Compute the relevance of each tailor\par
\noindent\hfill
$\displaystyle
s_i
\gets
\frac{1}{|\mathcal{C}_{\mathrm{tgt}}|}
\sum_{c\in\mathcal{C}_{\mathrm{tgt}}}
\left|
\left[\boldsymbol{\sigma}_{\mathcal{T}}^{c}\right]_i
\right|,
\qquad
i=1,\ldots,N_T
\mathpunct{;}$
\hfill\null\par

Construct the top-$K$ deployment mask\par
\noindent\hfill
$\displaystyle
m_K^{(i)}
\gets
\mathbb{I}\!\left[
i\in
\operatorname{TopK}
\left(
\{s_j\}_{j=1}^{N_T}
\right)
\right],
\qquad
i=1,\ldots,N_T
\mathpunct{;}$
\hfill\null\par

Select the target-relevant tailors\par
\noindent\hfill
$\displaystyle
\mathcal{T}_{\mathrm{tgt}}^{K}
\gets
\left\{
\Theta_{\star}^{(l,N_G+i)}
\in\mathcal{T}
\;\middle|\;
m_K^{(i)}=1
\right\}
\mathpunct{;}$
\hfill\null\par

Initialize the target model with
$\Theta_{\mathrm{tgt}}
\gets
\mathcal{G}\cup\mathcal{T}_{\mathrm{tgt}}^{K}$\;

Freeze $\mathcal{G}$, $\mathcal{H}$, and all unselected tailors\;

\BlankLine
\textbf{Selective Downstream Adaptation}\;

\While{not converged}{
    Sample an image--condition pair
    $(x,c)\sim\mathcal{D}_{\mathrm{tgt}}$\;

    Compute the condition representation
    $\mathbf{e}_c
    \gets
    \mathcal{E}_{\mathrm{cond}}(c)$\;

    Predict the tailor modulation values
    $\boldsymbol{\sigma}_{\mathcal{T}}^{c}
    \gets
    \mathcal{H}(\mathbf{e}_c)$\;

    \For{$l=1,\ldots,L$}{
        \ForEach{$\star\in\mathcal{S}$}{
            Reconstruct the target-specific weight

            \noindent\hfill
            $\displaystyle
            \widetilde{W}_{\star,c}^{(l)}
            \gets
            \mathcal{G}_{\star}^{(l)}
            +
            \sum_{i=1}^{N_T}
            m_K^{(i)}
            \left[\boldsymbol{\sigma}_{\mathcal{T}}^{c}\right]_i
            \mathcal{T}_{\star,i}^{(l)}
            \mathpunct{;}$
            \hfill\null\par
        }
    }

    Sample $t$ and
    $\boldsymbol{\epsilon}\sim\mathcal{N}(0,I)$, and construct
    $\mathbf{z}_t$\;

    Compute the diffusion objective
    \makebox[\linewidth][c]{%
        $\displaystyle
        \mathcal{L}_{\mathrm{diff}}
        \gets
        \left\|
        \boldsymbol{\epsilon}
        -
        \boldsymbol{\epsilon}_{\widetilde{\Theta}_{\mathrm{tgt}}}
        (\mathbf{z}_t\mid c,t)
        \right\|_2^2
        \mathpunct{;}$
    }

    Update only $\mathcal{T}_{\mathrm{tgt}}^{K}$
    by minimizing $\mathcal{L}_{\mathrm{diff}}$\;
}

\Return{$\mathcal{M}_{\mathrm{tgt}}
\left(\Theta_{\mathrm{tgt}}\right)$}\;
\label{alg:downstream}
\end{algorithm}

\subsection{Target-Aware Model Recombination}
\label{sec:app_recombination}

The decomposable parameterization learned by KDiv enables a target model to be assembled from the condition-agnostic learngenes and a compact subset of relevant tailors. The complete recombination and adaptation procedure is summarized in Algorithm~\ref{alg:downstream}.

Given a target dataset $\mathcal{D}_{\mathrm{tgt}}$, we denote its condition set by $\mathcal{C}_{\mathrm{tgt}}$. For each condition $c\in\mathcal{C}_{\mathrm{tgt}}$, the pretrained condition encoder and gating network produce the tailor modulation vector
$\boldsymbol{\sigma}_{\mathcal{T}}^{c}
=
\mathcal{H}(\mathcal{E}_{\mathrm{cond}}(c))$.
We estimate the relevance of the $i$-th tailor to the target distribution by averaging its modulation magnitude across the target conditions:
\begin{equation*}
s_i(\mathcal{C}_{\mathrm{tgt}})
=
\frac{1}{|\mathcal{C}_{\mathrm{tgt}}|}
\sum_{c\in\mathcal{C}_{\mathrm{tgt}}}
\left|
\left[\boldsymbol{\sigma}_{\mathcal{T}}^{c}\right]_i
\right|,
\qquad
i=1,\ldots,N_T.
\end{equation*}
Because the modulation values are real-valued, their absolute values measure the contribution strengths of the corresponding tailors irrespective of direction. When $\mathcal{C}_{\mathrm{tgt}}$ is large, the relevance scores can be efficiently estimated from a representative subset of target conditions.

The $K$ tailors with the highest relevance scores are retained using the binary mask
\begin{equation*}
m_K^{(i)}
=
\mathbb{I}\!\left[
i\in
\operatorname{TopK}
\left(
\{s_j(\mathcal{C}_{\mathrm{tgt}})\}_{j=1}^{N_T}
\right)
\right],
\qquad
i=1,\ldots,N_T,
\end{equation*}
where $\mathbb{I}[\cdot]$ denotes the indicator function. The same mask is applied across the decomposed projection matrices, yielding the selected tailor collection
\begin{equation*}
\mathcal{T}_{\mathrm{tgt}}^{K}
=
\left\{
\Theta_{\star}^{(l,N_G+i)}
\in\mathcal{T}
\;\middle|\;
m_K^{(i)}=1,\;
l\in[1,L],\;
\star\in\mathcal{S}
\right\}.
\end{equation*}
The target-model parameters are then initialized by combining the complete learngene collection with the selected tailors:
\begin{equation*}
\Theta_{\mathrm{tgt}}
=
\mathcal{G}
\cup
\mathcal{T}_{\mathrm{tgt}}^{K}.
\end{equation*}
This construction preserves the shared generative knowledge encoded by $\mathcal{G}$ while excluding tailors that contribute weakly to the target condition distribution.

For an input condition $c$, each target-specific projection matrix is reconstructed as
\begin{equation*}
\widetilde{W}_{\star,c}^{(l)}
=
\mathcal{G}_{\star}^{(l)}
+
\sum_{i=1}^{N_T}
m_K^{(i)}
\left[\boldsymbol{\sigma}_{\mathcal{T}}^{c}\right]_i
\mathcal{T}_{\star,i}^{(l)}.
\end{equation*}
The binary mask determines which tailors are retained in the target model, whereas the continuous modulation values control their condition-dependent contributions. KDiv therefore preserves fine-grained routing among the selected tailors rather than assigning each target condition to a single component.

After recombination, tailors excluded by the top-$K$ mask are discarded rather than retained as frozen parameters. The resulting target model therefore contains only the complete learngene collection $\mathcal{G}$ and the selected tailor collection $\mathcal{T}_{\mathrm{tgt}}^{K}$. During downstream adaptation, the learngenes and routing modules remain fixed, while only $\mathcal{T}_{\mathrm{tgt}}^{K}$ is optimized:
\begin{equation*}
\min_{\mathcal{T}_{\mathrm{tgt}}^{K}}
\;
\mathbb{E}_{(x,c)\sim\mathcal{D}_{\mathrm{tgt}}}
\left[
\mathcal{L}_{\mathrm{diff}}
\left(
\mathcal{E}(x),c;
\mathcal{G},
\mathcal{T}_{\mathrm{tgt}}^{K},
\mathcal{H}
\right)
\right],
\qquad
\nabla_{\mathcal{G}}\mathcal{L}_{\mathrm{diff}}
=
\nabla_{\mathcal{H}}\mathcal{L}_{\mathrm{diff}}
=
0.
\end{equation*}
This selective adaptation preserves the transferable generative prior encoded by $\mathcal{G}$ while concentrating optimization on target-relevant variations. The selection size $K$ controls both adaptation capacity and model size: a larger $K$ retains more specialized knowledge, whereas a smaller $K$ yields a more compact model with fewer active and trainable parameters. 
See Sec.~\ref{sec:tailor_rank} for an empirical analysis of the trade-off between tailor capacity, adaptation performance, and parameter efficiency.

\section{Additional Training Details}
\label{sec:app_training}

\subsection{Details of Downstream Tasks}
\label{sec:app_dataset}

We evaluate KDiv across three conditional generation settings, covering categorical, textual, and visual conditions. These settings assess the transferability of the learned knowledge across diverse domains and conditioning modalities. Table~\ref{tab:dataset_statistics} summarizes the datasets and their publicly reported sizes.

\subsubsection{\textsc{Class-Conditional Generation (C2I)}
}
We pre-train KDiv on ImageNet-1K~\cite{deng2009imagenet}, which contains 1,000 object categories, and evaluate downstream transfer on eight datasets spanning fine-grained recognition, food, natural scenes, and remote sensing. These datasets include CUB-200-2011~\citesupp{supp_wah2011caltech}, Danish Fungi 2020--Mini (DF20M)~\citesupp{supp_picek2022danish}, Flowers-102~\citesupp{supp_nilsback2008automated}, Food-101~\citesupp{supp_bossard2014food}, LSUN Bedroom and LSUN Church~\citesupp{supp_wang2017knowledge}, NWPU-RESISC45~\citesupp{supp_cheng2017remote}, and Stanford Cars~\citesupp{supp_gebru2017fine}. Together, these datasets cover substantial variation in object appearance, fine-grained categories, and scene characteristics, enabling us to assess transfer across heterogeneous image domains.

\subsubsection{\textsc{Text-Conditional Generation (T2I)}
}
We initialize from a pre-trained PixArt-$\alpha$ model~\cite{chen2024pixart} and continue pre-training on Fine-T2I~\cite{ma2026finet2i}, a large-scale dataset containing over six million image--text pairs. During this stage, KDiv reorganizes the model into decomposable learngene and tailor modules to facilitate the selective reuse of learned knowledge. We evaluate downstream transfer on eight datasets covering diverse visual domains: Adventurers~\citesupp{supp_rvorias2023realms}, FluentEmoji~\citesupp{supp_adler2023fluentemoji}, Hearthstone~\citesupp{supp_adler2023hearthstone}, LEGO~\citesupp{supp_adler2022lego}, Lunara~\citesupp{supp_wang2026moonworks}, Pokémon~\citesupp{supp_diffusers2023pokemongpt4}, RickMorty~\citesupp{supp_adler2022rickandmorty}, and Signatures~\citesupp{supp_jadekai2026signatures}. These datasets span fantasy characters, stylized graphics, game-related imagery, aesthetic styles, cartoon content, and handwritten signatures, testing the model's ability to transfer knowledge across distinct semantic and visual domains.

\subsubsection{\textsc{Image-Conditional Generation (I2I)}
}
We perform knowledge diversion on Subjects200K~\cite{tan2024ominicontrol}, a dataset containing approximately 200,000 paired images, using eight base conditions. We then evaluate transfer following CtrLoRA~\cite{xu2024ctrlora} on the COCO2017 validation set~\cite{lin2014microsoft}, which contains 5,000 images. The evaluation considers eight novel control types: Blur, Brush, Grayscale, Inpainting, JPEG compression, Palette, Pixel, and Shuffle. This setting evaluates whether knowledge organized under the base conditions can be recombined to support previously unseen visual transformations.

Overall, these datasets provide a diverse benchmark for evaluating generation quality and knowledge transfer across conditioning modalities, semantic domains, and visual transformations.

\begin{table*}[t]
\centering
\caption{
\textbf{Datasets used for pre-training and downstream adaptation.}
}
\vspace{-0.1in}
\label{tab:dataset_statistics}
\small
\setlength{\tabcolsep}{2pt}
\begin{tabularx}{\textwidth}{
@{}
>{\raggedright\arraybackslash}p{0.23\textwidth}
>{\raggedright\arraybackslash}p{0.20\textwidth}
>{\raggedright\arraybackslash}X
>{\raggedright\arraybackslash}p{0.25\textwidth}
@{}
}
\toprule[1pt]
\textbf{Dataset} & \textbf{Usage} & \textbf{Domain} & \textbf{Number of Samples} \\
\midrule[1pt]

\multicolumn{3}{@{}l}{
\textbf{\textsc{Class-Conditional Generation (C2I)}}
} \\
\midrule
ImageNet-1K~\cite{deng2009imagenet}
& Source Pre-training 
& over 1,000 object categories
& 1,281,167 train / 50,000 val \\

CUB-200-2011~\citesupp{supp_wah2011caltech}
& Downstream Adaptation 
& 200 fine-grained bird categories
& 11,788 images \\

DF20M~\citesupp{supp_picek2022danish}
& Downstream Adaptation 
& 182 fungal species from six genera
& 36,393 images \\

Flowers-102~\citesupp{supp_nilsback2008automated}
& Downstream Adaptation 
& 102 flower categories
& 8,189 images \\

Food-101~\citesupp{supp_bossard2014food}
& Downstream Adaptation 
& 101 food categories
& 101,000 images \\

LSUN Bedroom~\citesupp{supp_wang2017knowledge}
& Downstream Adaptation 
& indoor bedroom scenes
& 3,033,042 train / 300 val \\

LSUN Church~\citesupp{supp_wang2017knowledge}
& Downstream Adaptation 
& outdoor church scenes
& 126,227 train / 300 val \\

NWPU-RESISC45~\citesupp{supp_cheng2017remote}
& Downstream Adaptation 
& 45 remote-sensing scene categories
& 31,500 images \\

Stanford Cars~\citesupp{supp_gebru2017fine}
& Downstream Adaptation 
& 196 fine-grained car categories
& 16,185 images \\

\bottomrule[1pt]
\toprule[1pt]
\multicolumn{3}{@{}l}{
\textbf{\textsc{Text-Conditional Generation (T2I)}}
} \\
\midrule
Fine-T2I~\cite{ma2026finet2i}
& Source Pre-training 
& diverse image--text data
& approximately 6.3M pairs \\

Adventurers~\citesupp{supp_rvorias2023realms}
& Downstream Adaptation 
& fantasy characters and environments
& 830 image--text pairs \\

FluentEmoji~\citesupp{supp_adler2023fluentemoji}
& Downstream Adaptation 
& stylized emoji imagery
& 7,564 image--text pairs \\

Hearthstone~\citesupp{supp_adler2023hearthstone}
& Downstream Adaptation 
& fantasy game-card imagery
& 2,952 image--text pairs \\

LEGO~\citesupp{supp_adler2022lego}
& Downstream Adaptation 
& LEGO-style imagery
& 2,511 image--text pairs \\

Lunara~\citesupp{supp_wang2026moonworks}
& Downstream Adaptation 
& aesthetic and stylized imagery
& 2,000 image--text pairs \\

Pokémon~\citesupp{supp_diffusers2023pokemongpt4}
& Downstream Adaptation 
& Pokémon character imagery
& 833 image--text pairs \\

RickMorty~\citesupp{supp_adler2022rickandmorty}
& Downstream Adaptation 
& Rick and Morty cartoon imagery
& 530 image--text pairs \\

Signatures~\citesupp{supp_jadekai2026signatures}
& Downstream Adaptation 
& handwritten signature generation
& 1,289 image--text pairs \\

\bottomrule[1pt]
\toprule[1pt]

\multicolumn{3}{@{}l}{
\textbf{\textsc{Image-Conditional Generation (I2I)}}
} \\
\midrule
Subjects200K~\cite{tan2024ominicontrol}
& Source Pre-training 
& eight base conditions
& approximately 200K image pairs \\

COCO2017 validation~\cite{lin2014microsoft}
& Downstream Adaptation 
& eight novel control types
& 5,000 source images \\

\bottomrule[1pt]
\end{tabularx}
\end{table*}

\subsection{Details of PEFT Baselines}
\label{sec:app_baseline}

We compare KDiv with representative parameter-efficient fine-tuning (PEFT) methods spanning spectral, orthogonal, and low-rank adaptation. These methods retain most pretrained parameters and introduce a restricted set of trainable parameters for downstream adaptation. For a controlled comparison, all methods use the same pretrained backbone, target datasets, diffusion objective, and adaptation schedule. Their ranks or method-specific dimensions are selected to maintain comparable trainable parameter budgets whenever possible. The evaluated baselines are summarized below.

\begin{itemize}
    \item \textbf{SVDiff}~\cite{han2023svdiff} adapts a pretrained model in its spectral parameter space. It decomposes each pretrained weight matrix through singular value decomposition and optimizes compact spectral shifts while keeping the corresponding singular vectors fixed. This design modifies the strength of existing spectral components without directly updating the dense weights.

    \item \textbf{OFT}~\cite{qiu2023controlling} learns orthogonal transformations of pretrained weight matrices. By constraining the transformation to preserve hyperspherical geometry, OFT adapts the model while limiting distortion of the pretrained representation. Its structured orthogonal parameterization provides a parameter-efficient alternative to unconstrained weight updates.

    \item \textbf{LoRA}~\cite{hulora} represents the downstream update to a frozen weight matrix $W$ as a low-rank residual,
    $\Delta W=BA$, where the rank of $BA$ is substantially smaller than the dimensions of $W$. Only the low-rank factors $A$ and $B$ are optimized during adaptation, making LoRA a standard baseline for evaluating trainable-parameter efficiency.

    \item \textbf{PiSSA}~\cite{meng2024pissa} initializes low-rank adaptation factors using the principal singular components of pretrained weights. The remaining components form a frozen residual, while the principal components are optimized for the downstream task. This initialization places the trainable parameters in dominant spectral directions and typically accelerates adaptation relative to random LoRA initialization.

    \item \textbf{LoHa}~\cite{yeh2024navigating} parameterizes each weight update as the Hadamard product of two low-rank matrices. This multiplicative construction increases the effective rank and expressive capacity of the update without proportionally increasing the number of trainable parameters.

    \item \textbf{DoRA}~\cite{liudora} decomposes pretrained weights into magnitude and directional components and applies low-rank adaptation primarily to the weight direction. Separating these components provides greater flexibility than directly applying a low-rank residual to the original weight matrix while retaining parameter-efficient optimization.
\end{itemize}

For all PEFT baselines, the pretrained backbone parameters remain frozen and only the method-specific adaptation parameters are optimized. Evaluation is conducted using identical data preprocessing, training steps, and task-specific metrics to isolate differences in downstream adaptation capability.

\subsection{Hyperparameters}
\label{sec:hyper}
Tables~\ref{tab:hyper_main} and~\ref{tab:hyper_down} detail the hyperparameter configurations for knowledge-diversion pretraining and downstream adaptation, respectively. 
Table~\ref{tab:hyper_main} presents the training and sampling settings of KDiv across C2I, T2I, and I2I, including the optimization configurations and the numbers of learngenes ($N_G$) and tailors ($N_T$). 
Table~\ref{tab:hyper_down} reports the task-level training settings for C2I, T2I, and I2I, together with the method-specific learning rates, ranks, and block sizes used for KDiv and the competing adaptation methods.

\begin{table*}[tb]
    \centering
    \caption{\textbf{Hyperparameter configurations for pre-training decomposable models with KDiv.}}
    \vspace{-0.05in}
    \setlength{\tabcolsep}{5 mm}
    \resizebox{\textwidth}{!}{
        \begin{tabular}{@{}lrrr@{}}
        \toprule[1pt]
        \textsc{\textbf{Training Settings}} & \textsc{\textbf{Class-Conditional Gen.}} & \textsc{\textbf{Text-Conditional Gen.}} & \textsc{\textbf{Image-Conditional Gen.}} \\
        \midrule[1pt]
        Optimizer & AdamW & AdamW & AdamW\\
        Learning rate & $1{\times}10^{-4}$ & $1{\times}10^{-4}$ & $1{\times}10^{-5}$\\
        Weight decay & 0 & 0.03 & 0.03\\
        Warmup steps & 0 & 100 & 0\\
        LR scheduler & Constant & Constant (400K) + cosine decay (100K) & Constant\\
        Batch size & 256 & 256 & 16\\
        Training steps & 400,000 & 500,000 & 450,000\\
        Image resolution & 256$\times$256 & 256$\times$256 & 512$\times$512\\
        VAE & sd-vae-ft-ema & sd-vae-ft-ema & sd-vae-ft-ema\\
        CFG scale & 1.5 & 1.0 & 4.5\\
        Sample steps & 250 & 20 & 25\\
        $N_G$ & 96 / 192 (DiT-S/-B) & 200 & 576\\
        $N_T$ & 288 / 576 (DiT-S/-B) & 952 & 576\\
        \bottomrule[1pt]
        \end{tabular}
        }
    \label{tab:hyper_main}
\end{table*}

\begin{table*}[t]
    \centering
    \caption{
    \textbf{Hyperparameter configurations for downstream adaptation.}
    Task-level training settings and method-specific configurations used to evaluate KDiv and competing PEFT methods across class-, text-, and image-conditioned generation.
    }
    \vspace{-0.05in}
    \setlength{\tabcolsep}{8pt}
    \resizebox{\linewidth}{!}{%
        \begin{tabular}{@{}clcccccc@{}}
            \toprule[1pt]
            \textbf{Level} & \textbf{Hyperparameter} & \multicolumn{2}{c}{\textsc{\textbf{Class-Conditional Gen.}}} & \multicolumn{2}{c}{\textsc{\textbf{Text-Conditional Gen.}}} & \multicolumn{2}{c}{\textsc{\textbf{Image-Conditional Gen.}}} \\
            \midrule
            \multirow{7}{*}{\rotatebox[origin=c]{90}{\textsc{Task Level}}} & Batch size & \multicolumn{2}{c}{256} & \multicolumn{2}{c}{32} & \multicolumn{2}{c}{4} \\
            & Training steps & \multicolumn{2}{c}{50{,}000} & \multicolumn{2}{c}{3{,}000} & \multicolumn{2}{c}{3{,}000} \\
            & Optimizer & \multicolumn{2}{c}{AdamW} & \multicolumn{2}{c}{AdamW} & \multicolumn{2}{c}{AdamW} \\
            & Weight decay & \multicolumn{2}{c}{0} & \multicolumn{2}{c}{0.03} & \multicolumn{2}{c}{0.03} \\
            & LR scheduler & \multicolumn{2}{c}{Constant} & \multicolumn{2}{c}{Constant} & \multicolumn{2}{c}{Constant} \\
            & Warmup steps & \multicolumn{2}{c}{0} & \multicolumn{2}{c}{0} & \multicolumn{2}{c}{0} \\
            & Image resolution & \multicolumn{2}{c}{$256\times256$} & \multicolumn{2}{c}{$256\times256$} & \multicolumn{2}{c}{$512\times512$} \\
            \bottomrule[1pt]
            \toprule[1pt]
            \textbf{Level} & \textbf{Method} & \textbf{LR} & \textbf{Rank/Block Size} & \textbf{LR} & \textbf{Rank/Block Size} & \textbf{LR} & \textbf{Rank/Block Size} \\
            \midrule
            \multirow{8}{*}{\rotatebox[origin=c]{90}{\textsc{Method Level}}} & SVDiff & $1{\times}10^{-4}$ & \textemdash & $1{\times}10^{-4}$ & \textemdash & $1.25{\times}10^{-3}$ & \textemdash \\
            & OFT & $1{\times}10^{-4}$ & 1536 & $1{\times}10^{-4}$ & 1152 & $6.25{\times}10^{-4}$ & 192 \\
            & LoRA & $1{\times}10^{-4}$ & 256 & $1{\times}10^{-3}$ & 200 & $1.25{\times}10^{-3}$ & 36 \\
            & PiSSA & $1{\times}10^{-4}$ & 256 & $1{\times}10^{-3}$ & 200 & $1.25{\times}10^{-3}$ & 36 \\
            & LoHa & $1{\times}10^{-4}$ & 128 & $1{\times}10^{-3}$ & 100 & $1.25{\times}10^{-3}$ & 18 \\
            & DoRA & $1{\times}10^{-4}$ & 256 & $1{\times}10^{-3}$ & 200 & $1.25{\times}10^{-3}$ & 36 \\
            & KIND & $1{\times}10^{-4}$ & 256 & $1{\times}10^{-3}$ & 200 & \textemdash & \textemdash \\
            & KDiv & $1{\times}10^{-4}$ & 256 & $1{\times}10^{-3}$ & 200 & $1.25{\times}10^{-3}$ & 36 \\
            \bottomrule[1pt]
        \end{tabular}%
    }
    \label{tab:hyper_down}
\end{table*}

\section{Additional Results}

\begin{figure*}[tb]
  \centering
  \includegraphics[width=\linewidth]{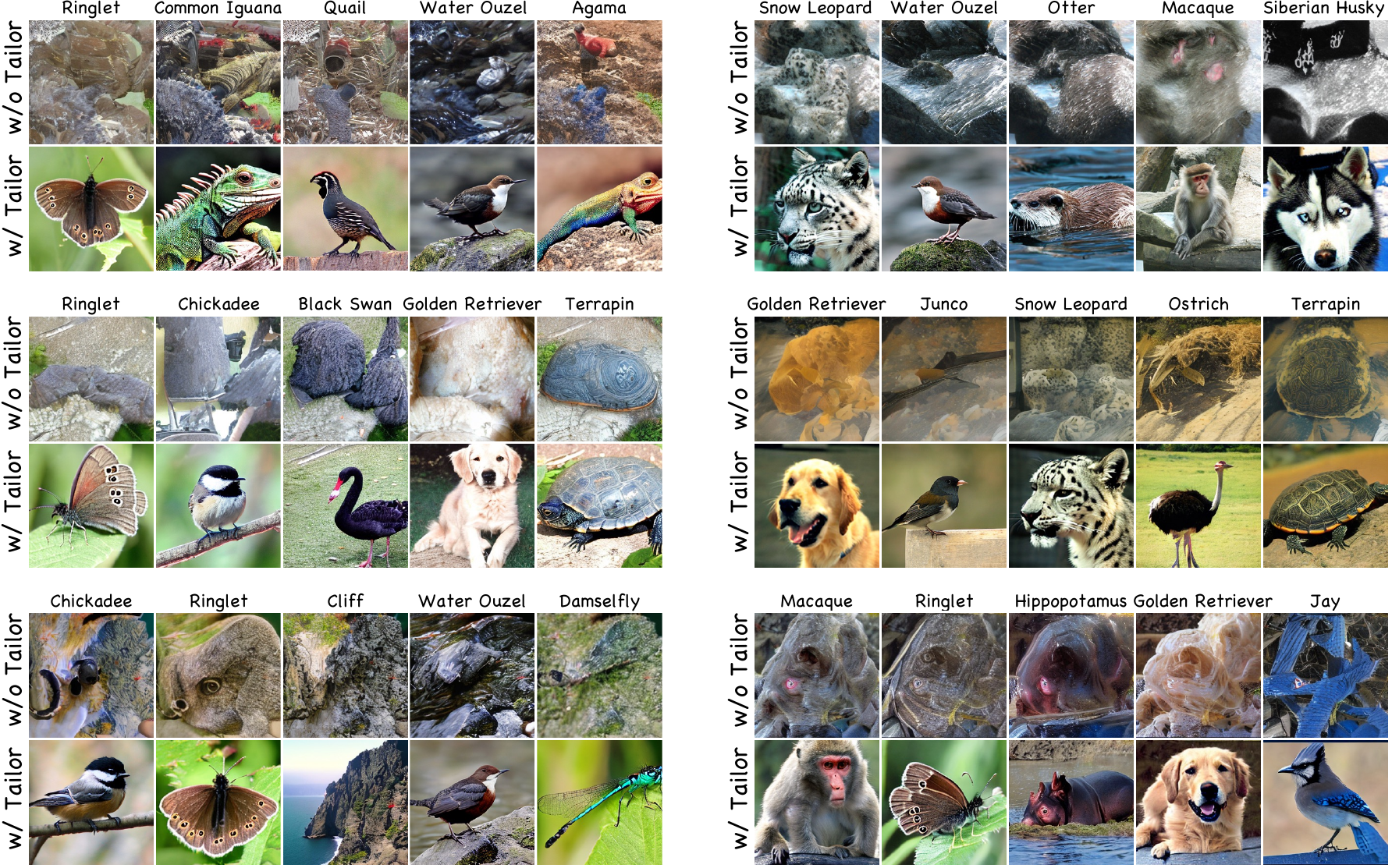}
  \vspace{-0.2in}
  \caption{\textbf{Additional visualization of condition-agnostic knowledge in learngenes.}
  Learngene-only outputs exhibit no condition-specific semantics, reflecting the condition-agnostic knowledge encoded by learngenes, whereas incorporating tailors introduces the semantics required for condition-aligned generation.}
  \label{fig:app_task-agnostic}
\end{figure*}

\subsection{Additional Analysis of Learngene Condition Invariance}

Fig.~\ref{fig:app_task-agnostic} further illustrates the functional separation between learngenes and tailors by varying class conditions and random seeds independently. 
Under a fixed random seed, learngene-only generation exhibits similar visual structures and appearance patterns across different class labels, indicating that learngenes encode little class-specific semantic information. 
Changing the random seed nevertheless produces diverse outputs, showing that this invariance is not caused by generative collapse. Instead, learngenes retain a diverse generative prior that remains largely independent of the input condition.

Incorporating the corresponding tailors aligns the generated samples with their specified class labels.
This comparison distinguishes stochastic variation from conditional semantics, suggesting that learngenes capture broadly reusable generative knowledge, while tailors provide the specialized information required for condition-aligned generation. The consistent results across different seeds and categories further support the systematic separation of condition-agnostic and condition-specific knowledge in KDiv.

\begin{figure*}[tb]
\centering
\includegraphics[width=\linewidth]{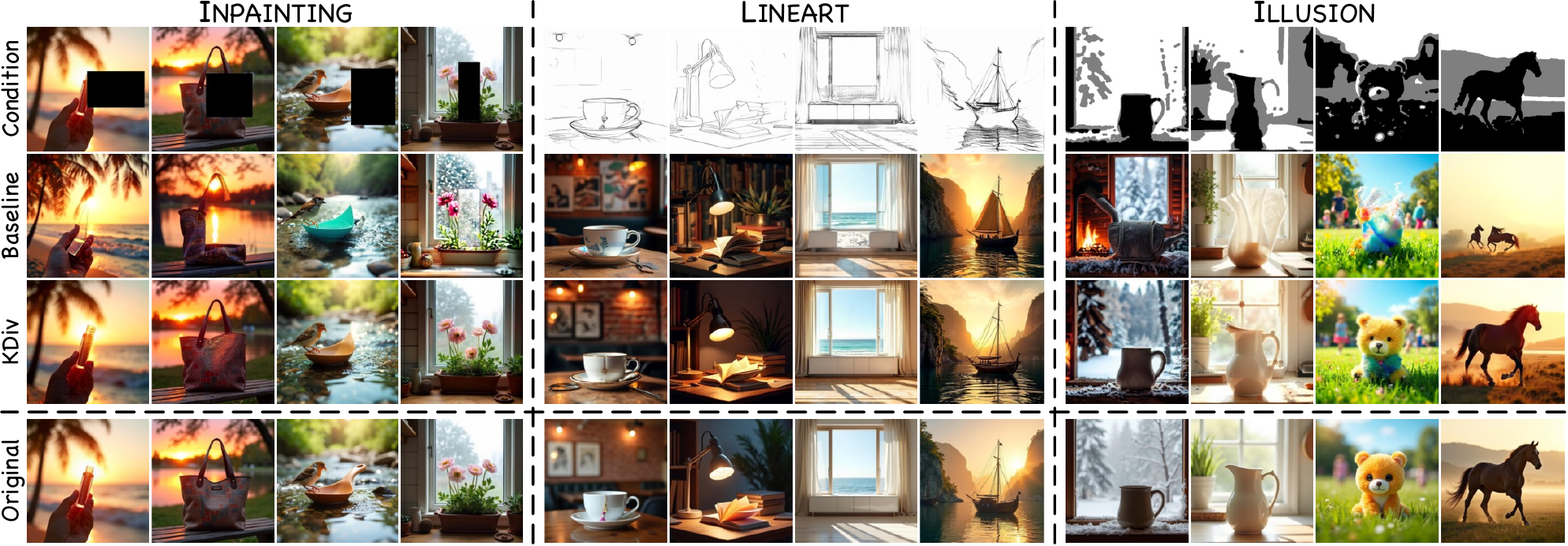}
\vspace{-0.2in}
\caption{
\textbf{Zero-shot generalization to unseen image conditions.}
Qualitative comparison of the dense baseline and KDiv under unseen \textsc{Inpainting}, \textsc{Lineart}, and \textsc{Illusion} conditions without downstream adaptation. KDiv demonstrates stronger condition adherence and better preserves the semantic content and spatial structure of the reference images.
}
\label{fig:app_i2i_zeroshot}
\end{figure*}

\subsection{Zero-Shot Generalization to Unseen Image Conditions}
\label{sec:app_zero}
Fig.~\ref{fig:app_i2i_zeroshot} evaluates the zero-shot generalization of KDiv to three image-conditioning types unseen during pre-training. Across inpainting masks, line drawings, and illusion-based structural cues, KDiv more faithfully preserves the semantic content and spatial organization of the reference images. It generates coherent content in masked regions, follows the geometry specified by line-art inputs, and recovers the global structures implied by illusion conditions.

Although the dense baseline produces visually plausible results, its outputs deviate more substantially from the input conditions, revealing limited robustness to shifts in conditioning structure. KDiv instead combines the transferable generative prior encoded by learngenes with relevant condition-specific knowledge from tailors. These results indicate that continuous condition-aware routing enables effective knowledge reuse for previously unseen controls, supporting zero-shot generalization without task-specific adaptation.

\begin{figure*}[tb]
\centering
\includegraphics[width=\linewidth]{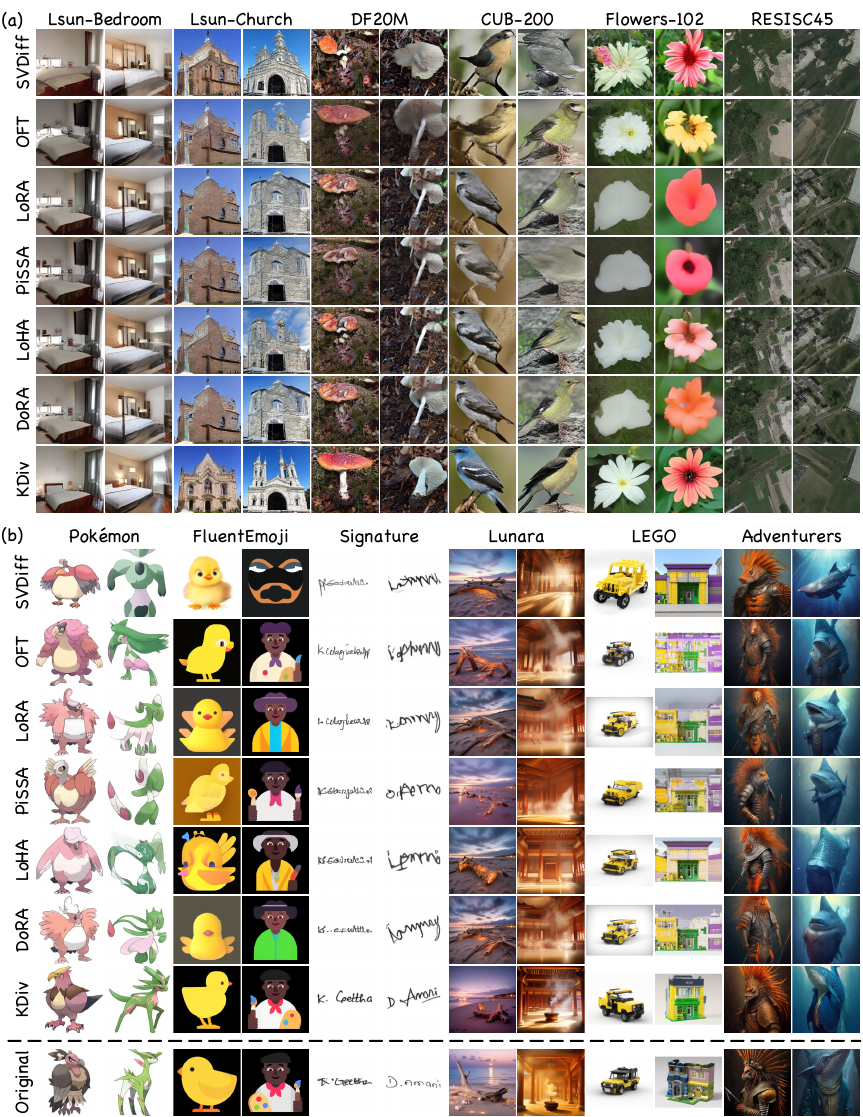}
\vspace{-0.2in}
\caption{
\textbf{Additional qualitative comparisons on downstream \textsc{Class-} and \textsc{Text-Conditional Generation} tasks.}
}
\label{fig:app_down_C2I_T2I}
\end{figure*}

\begin{figure*}[tb]
\centering
\includegraphics[width=\linewidth]{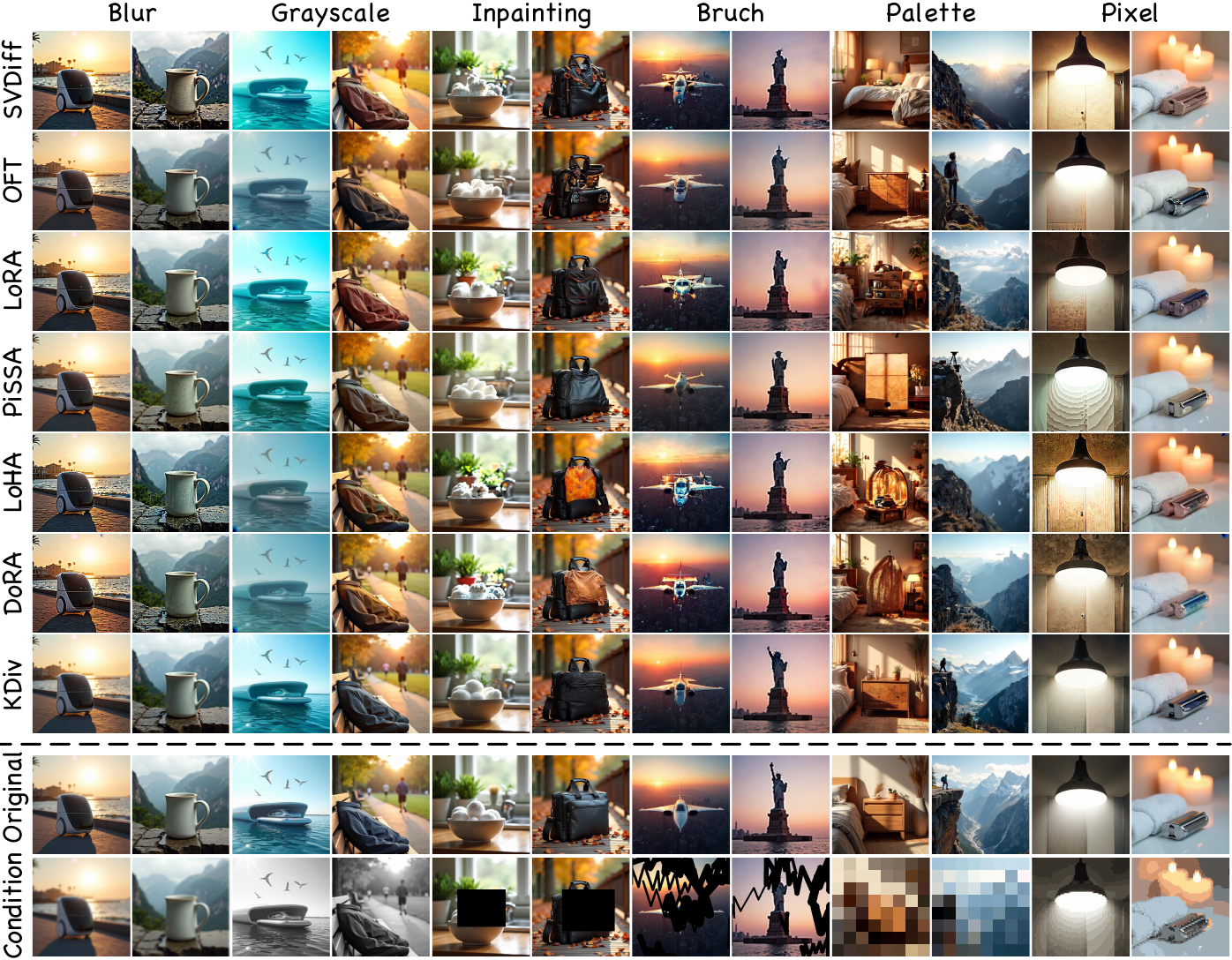}
\vspace{-0.2in}
\caption{
\textbf{Additional qualitative comparisons on downstream \textsc{Image-Conditional Generation} tasks.}
}
\label{fig:app_down_I2I}
\end{figure*}

\begin{figure*}[tb]
\centering
\includegraphics[width=\linewidth]{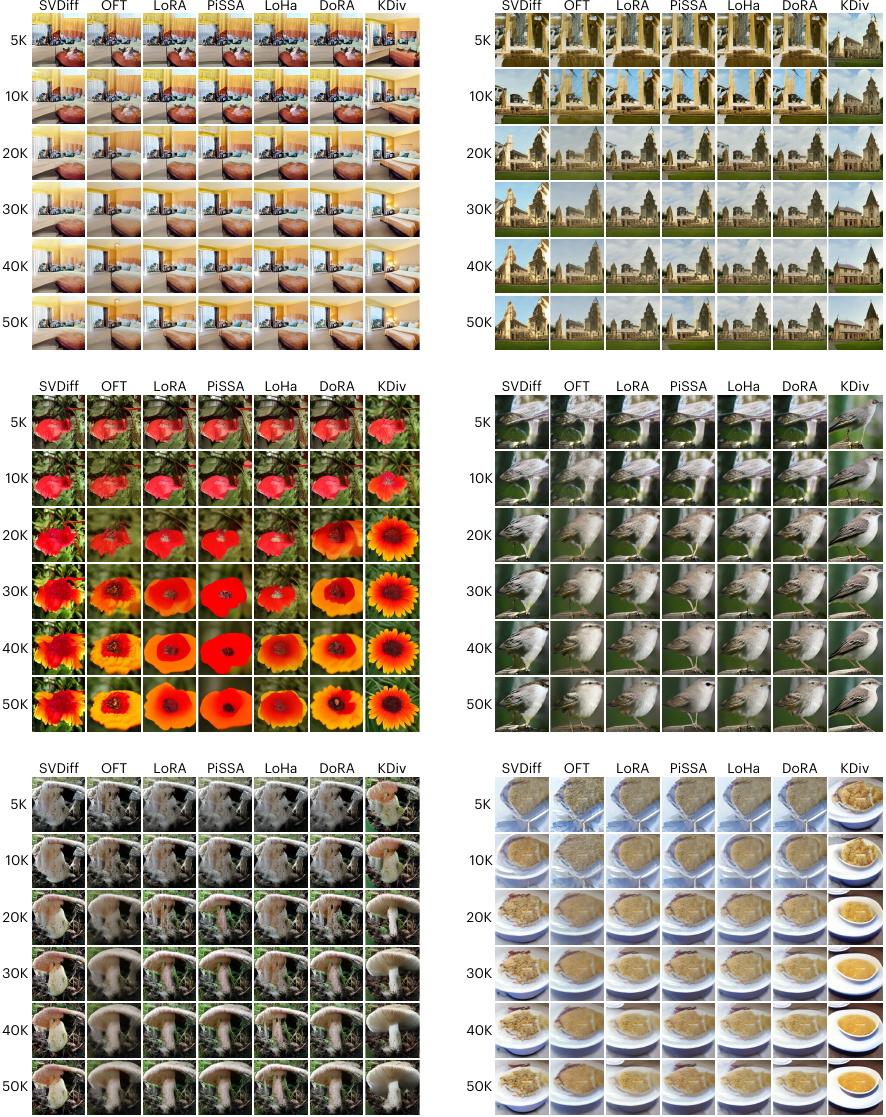}
\vspace{-0.2in}
\caption{
\textbf{Additional qualitative evolution of \textsc{Class-Conditional Generation (C2I)} samples at 10K-step training intervals.}
}
\label{fig:app_learn}
\end{figure*}

\section{Limitations and Future Work}

Despite its effectiveness across diverse conditional generation tasks, KDiv presents several limitations that motivate future research:

\begin{itemize}
    \item \textbf{Predefined Knowledge Allocation.}
    KDiv currently requires predefined numbers of learngenes $N_G$ and tailors $N_T$, as well as a downstream selection size $K$, whose optimal values may vary across architectures and tasks. 
    Future work could explore data-driven allocation and adaptive selection to reduce manual configuration.

    \item \textbf{Pre-training Overhead.}
    Factorized optimization, condition-aware routing, and dynamic weight reconstruction incur additional computational and memory costs during pre-training. Future work could mitigate this overhead through sparse routing, weight caching, and fused reconstruction.

    \item \textbf{Scope of Evaluation.}
    Our evaluation focuses on class-, text-, and image-conditioned generation. 
    Extending KDiv to text generation and discriminative tasks would further establish its generality across architectures and learning paradigms.
\end{itemize}

\clearpage
\bibliographystylesupp{IEEEtran}
\bibliographysupp{reference_supp.bib}

\end{document}